\documentclass[preprint,3p,twocolumn]{elsarticle}

\usepackage{amssymb}
\usepackage{amsmath}
\usepackage{bbm}

\usepackage{hyperref}
\usepackage{multirow}
\usepackage{soul}
\usepackage{todonotes}
\usepackage{pgfplots}
\usepackage{subcaption}
\usepackage{float}
\usepackage{numprint}
\usepackage{pifont}
\usepackage{pgfplotstable}
\usetikzlibrary{patterns}
\usepackage{arydshln}
\usepackage{diagbox}
\usepackage{booktabs}

\usepgfplotslibrary{groupplots} 
\usepgfplotslibrary{fillbetween}
\pgfplotsset{compat=1.18}

\newsavebox\CBox
\def\textBF#1{\sbox\CBox{#1}\resizebox{\wd\CBox}{\ht\CBox}{\textbf{#1}}}

\begin{document}

\begin{frontmatter}



\title{From Spherical to Gaussian: A Comparative Analysis of Point Cloud Cropping Strategies in Large-Scale 3D Environments}


\author[label1]{Maximilian Kellner}
\author[label1]{Dominik Merkle}
\author[label1,label2]{Michael Brunklaus}
\author[label1,label2]{Alexander Reiterer}

\affiliation[label1]{organization={Fraunhofer Institute for Physical Measurement Techniques IPM},
             city={Freiburg},
             postcode={79110},
             country={Germany}}

\affiliation[label2]{organization={University of Freiburg, Department of Sustainable Systems Engineering INATECH},
             city={Freiburg},
             postcode={79110},
             country={Germany}}

\begin{abstract}
Large-scale 3D point clouds can consist of hundreds of millions of points. Even after downsampling, these point clouds are too large for modern 3D neural networks. In order to develop a semantic understanding of the scene, the point clouds are divided into smaller subclouds that can be processed. Typically, this division is done using spherical crops, resulting in a loss of surrounding geometric context. To address this issue, we propose alternative methods that produce subclouds with larger crop sizes while maintaining a similar number of points. Specifically, we compare exponential, Gaussian, and linear cropping methods with the spherical method. We evaluated three 3D deep learning model architectures using multiple indoor and outdoor environment datasets. Our results demonstrate that altering the cropping strategy can enhance model performance, especially for large-scale outdoor scenes, yielding new state-of-the-art results. Code is available at \url{https://github.com/mvg-inatech/point_cloud_cropping}
\end{abstract}




\end{frontmatter}


\section{Introduction}\label{ch:introduction}
Recent advancements in 3D data acquisition have generated significant interest in analyzing point clouds. A variety of approaches have been developed to address this issue. The employment of diverse representations, encompassing 2D projections and discretization methods, has been instrumental in propelling the advancement of research in this domain. Pioneering works such as the deep learning model PointNet~\cite{qi_pointnet_2016} have emerged, offering innovative approaches for directly processing point clouds. 

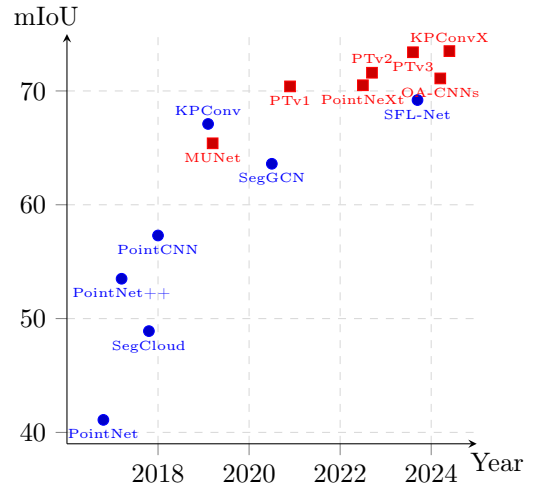
\begin{figure}[ht!]
    \centering
    \begin{tikzpicture}
    \begin{axis}[
        axis lines = middle,
        xlabel = {Year},
        ylabel = {mIoU},
        width=7cm,
        height=7cm,
        xlabel style={at={(axis description cs:1.05,0)}, anchor=north},
        ylabel style={at={(axis description cs:-0.05,1)}, anchor=south},
        /pgf/number format/1000 sep={},
        grid = major,
        grid style={dashed, gray!30},
        xmin = 2016, xmax = 2025,
        ymin = 39, ymax = 75,
        tick align=outside,
    ]

    \addplot+[
        only marks,
        nodes near coords,
        point meta=explicit symbolic,
        visualization depends on={value \thisrow{anchor}\as\myanchor},
        every node near coord/.append style={anchor=\myanchor, font=\tiny}
    ] table[col sep=comma, meta=label] {data/data_blue.csv};

    \addplot+[
        only marks,
        nodes near coords,
        point meta=explicit symbolic,
        visualization depends on={value \thisrow{anchor}\as\myanchor},
        every node near coord/.append style={anchor=\myanchor, font=\tiny}
    ] table[col sep=comma, meta=label] {data/data_red.csv};
        
    \end{axis}
    \end{tikzpicture}
    \caption{Different architectures tested on S3DIS \cite{armeni_joint_2017} Area 5. Blue indicates work that uses input point cloud cropping without prior knowledge of room boundary annotations. Red indicates the precise cutting of rooms, which simplifies the segmentation task.}
    \label{fig:rooms_vs_full_scene}
\end{figure}

Nonetheless, there exist challenges in deep learning that are unique to point clouds. One such challenge relates to the feasibility of processing large-scale scenes. 3D sensors have the capacity to scan distances of hundreds of meters. Furthermore, scenes can be captured from multiple scanning positions and fused to a single point cloud. The processing of these large-scale point clouds using deep learning is not feasible in their entire size due to GPU memory constraints, even with the implementation of novel methodologies and the utilization of advanced computational hardware.

To enable deep learning on large point clouds, the point cloud must be divided into smaller parts or downsampled in order to reduce the number of points. The resulting smaller point clouds, called subclouds in the following, can then be processed individually. Current state-of-the-art (SOTA) deep learning pipelines typically partition the point cloud into small overlapping spheres or cubes, causing loss of environmental context.

Consider the S3DIS \cite{armeni_3d_2016, armeni_joint_2017} dataset as an indoor example. The dataset comprises six point clouds of large areas. However, the dataset is also available as point clouds split into individual rooms. This allows to either load entire areas and calculate subcloud crops from them, which are then processed individually, or process individual rooms directly, as these are small enough to fit into a single mini-batch. In \cite{qian_pointnext_2022}, it was indicated that using full rooms instead of sampled subclouds would lead to improvement. This was confirmed in \cite{thomas_kpconvx_2024}. In Figure~\ref{fig:rooms_vs_full_scene} the most common architectures by their year of publication and performance are visualized using the mean Intersection over Union (mIoU) metric. Please note that only models that were trained exclusively on the dataset are included. Blue methods compute subclouds of the large scene, while red methods use single rooms as input, optionally dropping some points, which simplifies the segmentation task. Examining the graph reveals that all new, high-performing methods follow the single-room approach.

The finding of \cite{qian_pointnext_2022, thomas_kpconvx_2024} that the model benefits from viewing an entirely enclosed scene raises the question of whether the improvement comes from the architectures itself or the way the point cloud is presented. An entirely closed scene can be achieved in indoor scenes by splitting the scene into rooms initially. However, this requires splitting the room beforehand and can become problematic for rooms that are too large, causing the model to run out of memory. The problem becomes even more difficult if the scene cannot be split easily into independent small parts, as is often the case with outdoor scenes.

This study will investigate different cropping strategies for large-scale 3D environments. We will evaluate common approaches as well as proposed numerical alternatives, such as Gaussian, exponential, and linear cropping. We will compare these methods using multiple large-scale indoor and outdoor scene datasets. Additionally, we will use three different established 3D deep learning models to study their ability to adapt to different cropping strategies. We will show that alternatives to simple spherical subcloud cropping can achieve comparable results for indoor scenes and improve performance for outdoor scenes. The selected cropping approaches do not introduce additional computational complexity and are easy to adapt. We demonstrate that SOTA results can be achieved by simply adapting the cropping strategy, especially for outdoor scenes. Our main contribution can be summarized as follows:

\begin{itemize}
    \item We propose a fast and easy alternative subcloud cropping methods to avoid clear-cuts and the loss of surrounding context.
    \item We perform extensive experiments to evaluate the methods against a variety of hyperparameters for different indoor and outdoor scenes.
    \item We improve the SOTA results, especially for outdoor scenes, by using the proposed alternative subcloud cropping methods.
\end{itemize}

\section{Related work}\label{ch:related_work}
\textbf{3D semantic segmentation.} It can be divided into four categories using different representations: These are projection-based, discretization-based, point-based, and hybrid methods \cite{guo_deep_2021, zhang_deep-learning-based_2023}. Projection-based methods convert 3D point clouds into 2D images. This can be spherical view \cite{milioto_rangenet_2019, bebis_salsanext_2020, wu_squeezeseg_2017, wu_squeezesegv2_2019}, top down view \cite{aksoy_salsanet_2019, zhang_polarnet_2020, avidan_pillarnet_2022} or multiple views \cite{gerdzhev_tornado-net_2021, chen_mvlidarnet_2020, alnaggar_multi_2020}. Discretizing the point cloud into voxels allows the use of 3D convolution \cite{cicek_3d_2016, tchapmi_segcloud_2017}. The introduction of sparse 3D convolution \cite{graham_3d_2018, choy_4d_2019, spconv_contributors_spconv_2022}  enabled the efficient processing of large discretized scenes. Due to the possibility of scaling 2D kernels \cite{ding_scaling_2022}, different methods are proposed to scale up sparse 3D kernels \cite{chen_largekernel3d_2023, feng_lsk3dnet_2024}. Alternatively, an adaptive receptive field is proposed in \cite{peng_oa-cnns_2024}. Directly processing the point cloud using special Multi-layer Perceptrons (MLPs) is done in \cite{qi_pointnet_2016}. The work is improved in \cite{qi_pointnet_2017} using a hierarchical encoding MLP structure and in \cite{qian_pointnext_2022} using an improved training strategy. Using random sampling and local feature aggregation with MLPs is done in \cite{hu_randla-net_2020}. Alternative convolutions are proposed using graph structures \cite{landrieu_large-scale_2018, wang_dynamic_2019} or fuzzy spherical kernels \cite{lei_seggcn_2020}. Tangent convolutions are used in \cite{tatarchenko_tangent_2018}. In \cite{wu_pointconv_2020}, a compromise between weight and density functions is proposed. KPConv \cite{thomas_kpconv_2019} defines kernel points located in Euclidean space, which gives more flexibility than fixed grid convolutions. Kernel points are made lightweight in \cite{li_sfl-net_2023} and further improved in \cite{thomas_kpconvx_2024}. Combining MLPs with attention mechanisms is proposed in \cite{zhao_point_2021, guo_pct_2021}. The work is improved by grouped vector attention in \cite{wu_point_2022} and further improved using serialized points in \cite{wu_point_2024}. The architecture is simplified in terms of parameters and improved in runtime in \cite{yue_litept_2026}. Alternatively, Swin transformers can be used similarly to \cite{liu_swin_2022} by employing sparse voxels \cite{yang_swin3d_2025}. Hybrid approaches combine multiple representations to reduce drawbacks. While multiple views are fused using kernel points in \cite{kellner_fused_2022}, points and voxel features are combined in \cite{hou_point--voxel_2022}. \cite{xu_rpvnet_2021} combines all three named representations.

Due to the increasing popularity of self-supervised learning in computer vision \cite{caron_emerging_2021, he_masked_2022, oquab_dinov2_2024, simeoni_dinov3_2025}, the research community has begun adapting these methods for 3D tasks. Contrastive adaptations, such as those described in \cite{xie_pointcontrast_2020}, use different views of the same object or scene to ensure similar representations in the latent space. Alternatively, masked autoencoders divide the point cloud into patches and mask some patches out. The model then learns to reconstruct the masked patches, as described in \cite{avidan_masked_2022}. Using contrastive learning with masked scenes is done in \cite{wu_masked_2023}. A self-distillation framework paired with a mixture of multiple datasets is proposed in \cite{wu_sonata_2025}. They use local-global view alignment as well as mask-unmask view alignment within the framework. This method combined with 2D-3D cross-modal joint embedding is proposed in \cite{zhang_concerto_2025} to follow the concept of  multi sensor synergy. An alternative multi-modal approach is proposed in \cite{zhu_ponderv2_2025}.

\textbf{3D point cloud cropping.} Also referred to point cloud clipping, involves dividing a point cloud into multiple smaller parts. This is necessary when the total size of the point cloud cannot be processed all at once. For the most common datasets, such as \cite{behley_semantickitti_2019, sun_scalability_2020, caesar_nuscenes_2020}, which are used for perception tasks in autonomous driving, this is achieved by working with a single sensor sweep at a given position. This allows for cropping a box around the position using the commonly used distances between 30-70 meters. To this end, the problem of cropping the point cloud is solved implicitly by the data recording setup. Using multi-frame training and a no-clipping-point policy with \cite{wu_point_2024} in \cite{sun_scalability_2020} improves the model by more than 2~\% in terms of mIoU. This shows that having more context improves performance. However, this split is not as easy with large-scale point clouds such as \cite{roynard_paris-lille-3d_2018, tan_toronto-3d_2020, kellner_semanticbridge_2025}. The most common approaches are either sampling a fixed number of neighborhood points \cite{hu_randla-net_2020} or spherically crop the point cloud into smaller parts \cite{thomas_semantic_2018}. In \cite{hu_randla-net_2020} approximately $10^5$ points are sampled, while the input radius for cropping in \cite{thomas_kpconv_2019} for outdoor data is about 3~m or 4~m for \cite{varney_pyramid_2020}. Some use the original height as an additional input feature to have at least some global context. More modern architectures allow for larger amount of points and larger input radii, however the crop results in a loss of environmental context. In \cite{yoo_human_2023} a segmentation network for point clouds inspired by human peripheral vision is proposed. They address the same issue about the limited environmental context of the input. They solve it by introducing a multi-scale input and a parallel processing network with connection blocks. It is further improved in \cite{yoo_eyenet_2025}. However, we want to decouple this prior step from the deep learning architecture to allow for greater flexibility in terms of both the cropping method and the model. Modern 3D models are capable of learning hierarchical structures by themselves, and this way, the cropping can easily be incorporated into future models.

Many methodologies have been proposed to address the issue of 3D point cloud semantic segmentation. However, the research regarding how large-scale point clouds can be cropped into small subclouds to allow training those models without loosing resolution or surrounding context is limited.

\section{Method}\label{ch:method}
\subsection{Problem formulation}

Given a large point cloud $P \in \mathbb{R}^{N \times 3}$ with $N$~points and its coordinates $(x_i,y_i,z_i)_{i=1}^{N}$, the computational requirements increase as the number of points, $N$, increases. This can be addressed by reducing the number of points overall using voxel or random downsampling, which decreases the resolution. Alternatively, the input point cloud can be split into a set of subclouds, $\mathcal{X}$. This creates a trade-off between the resolution of the model and the surrounding context that the model can see.

The most commonly used method to split the scene is spherical cropping. A center point $\mathbf{x}_c$ is chosen with a radius $r$ resulting in a subcloud $\mathcal{X}^p = \{ \mathbf{x}_i \in P \,|\, \| \mathbf{x}_i - \mathbf{x}_c \| \leq r \} $. An alternative is the cube with an edge length $a$, resulting in $ \mathcal{X}^p = \{ \mathbf{x}_i \in P \,|\, |x_i^x - x_c^x| \leq a, \, |x_i^y - x_c^y| \leq a, \, |x_i^z - x_c^z| \leq a \} $. However, both cropping strategies result in a clear cutoff within the full point cloud, causing a loss of environmental context. This loss of surrounding context limits the model's performance.

\begin{figure}
    \centering
    \begin{tikzpicture}
    \begin{axis}[
        axis lines = middle,
        xlabel = $d$,
        ylabel = $p$,
        xlabel style={at={(axis description cs:1,0)}, anchor=north},
        ylabel style={at={(axis description cs:-0.05,1)}, anchor=south},
        grid = major,
        grid style={dashed,gray!30},
        domain = 0:10,
        samples = 200,
        xmin = 0, xmax = 11,
        ymin = 0, ymax = 1.1,
        tick align=outside,
        legend style={
            at={(0.5,-0.15)},     
            anchor=north,         
            legend columns=2,     
        },
    ]

    \addplot[blue!70!black, ultra thick] coordinates {
        (0,1) (5,1) (5,0) (10,0)
    };
    \addlegendentry{$p_s$}
    
    \addplot[green!70!black, ultra thick, smooth] {exp(-0.4*x)};
    \addlegendentry{$p_e$}
    
    \addplot[red!70!black, ultra thick, smooth] {exp(-((x/4)^2))};
    \addlegendentry{$p_g$}
    
    \addplot[black, ultra thick, smooth] {(10-x)/10};
    \addlegendentry{$p_l$}


    
    \end{axis}
    \end{tikzpicture}
    \caption{Probability of a point being selected depending on its distance to a selected center point.}
    \label{fig:distributions_over_distance}
\end{figure}
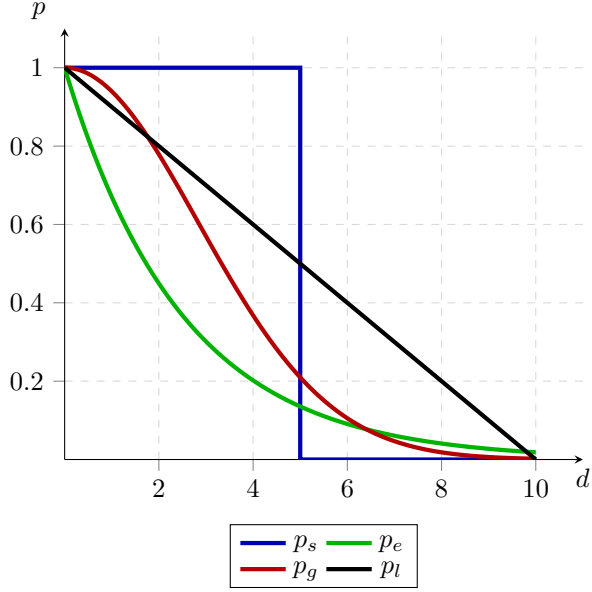

\subsection{Alternative cropping}

Instead of using this clear-cut method, we propose using different distributions to determine the probability $p$ that a point will be selected, given a center $\mathbf{x}_c$ and the euclidean distances between each point $\mathbf{x}_i$ and the center $d_i = \|\mathbf{x}_i - \mathbf{x}_c\|, \quad i = 1, 2, \ldots, N$. In this study, we propose exponential, Gaussian, and linear cropping and compare it to standard spherical cropping using the following probability equations:

\begin{align}
    p_s &= 
        \begin{cases} 1 & \text{if } d < d_m \\ 0 & \text{if } d \geq d_m \end{cases} \\
    p_e &= e^{-\lambda d} \\
    p_g &= e^{-(\frac{d}{\sigma_d})^2} \\
    p_l &= \frac{d_m -d}{d_m}
\end{align}

\begin{figure}
    \centering
    \begin{subfigure}[c]{0.45\textwidth}
        \includegraphics[width=\textwidth]{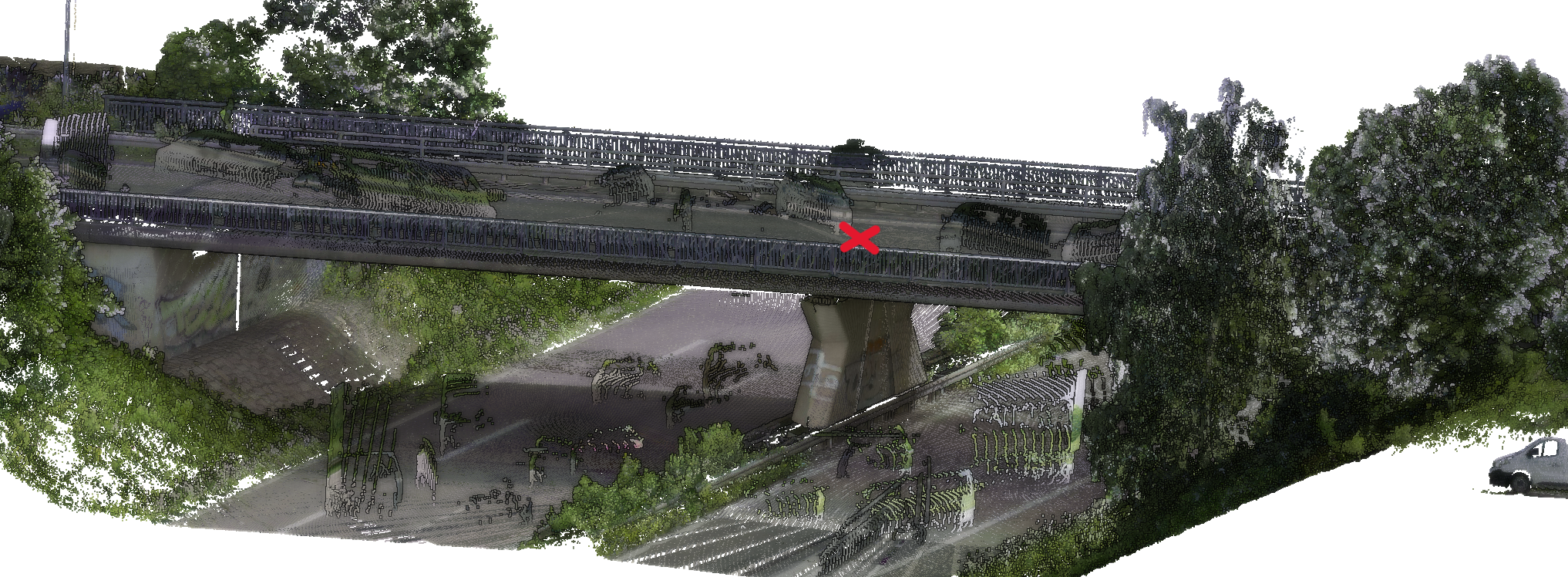}
        \caption{Full point cloud}
        \label{fig:point_cloud_crops_a}
    \end{subfigure}
    \begin{subfigure}[c]{0.22\textwidth}
        \includegraphics[width=\textwidth]{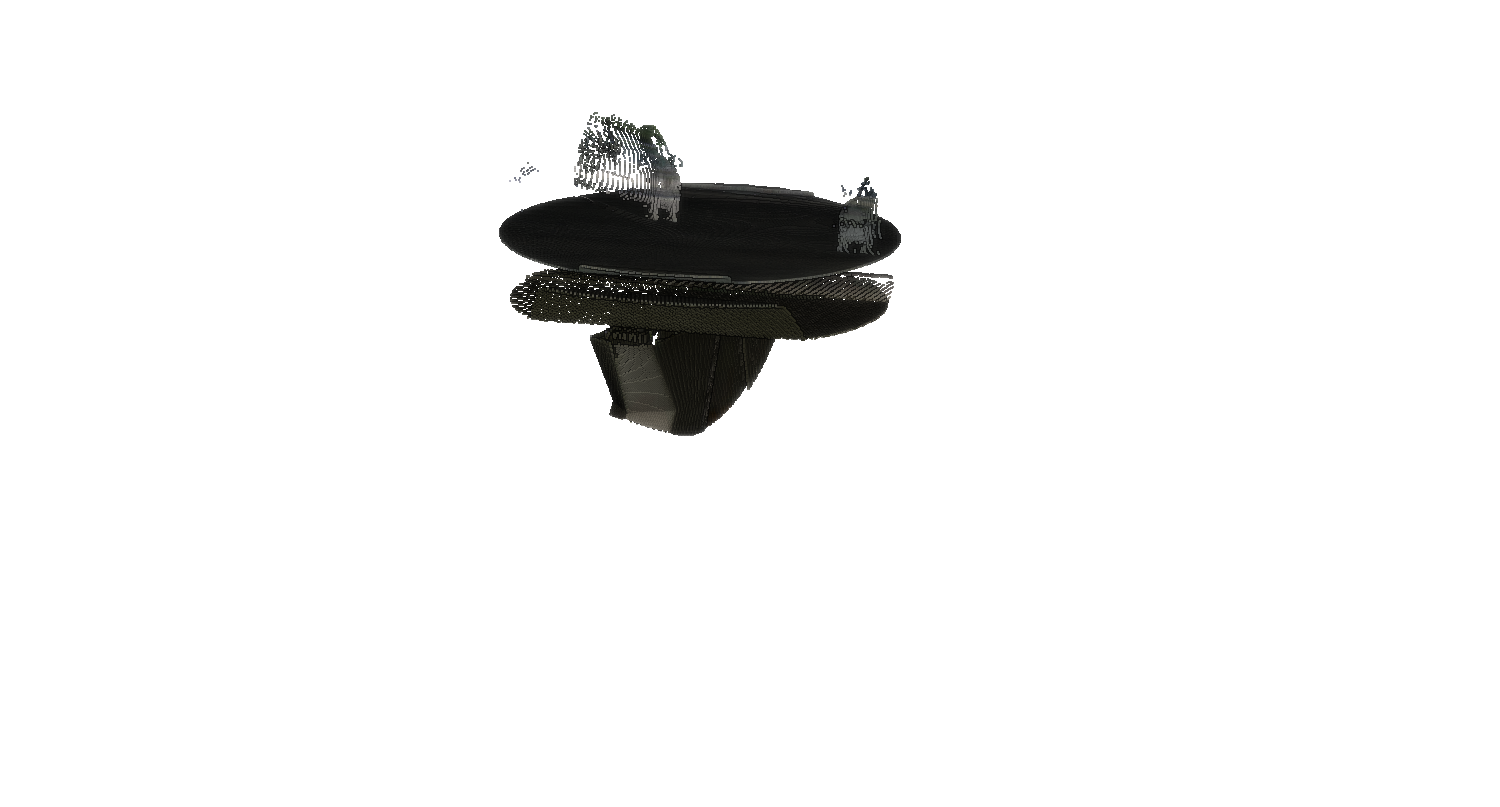}
        \caption{Spherical crop $p_s$}
    \end{subfigure}
    \begin{subfigure}[c]{0.22\textwidth}
        \includegraphics[width=\textwidth]{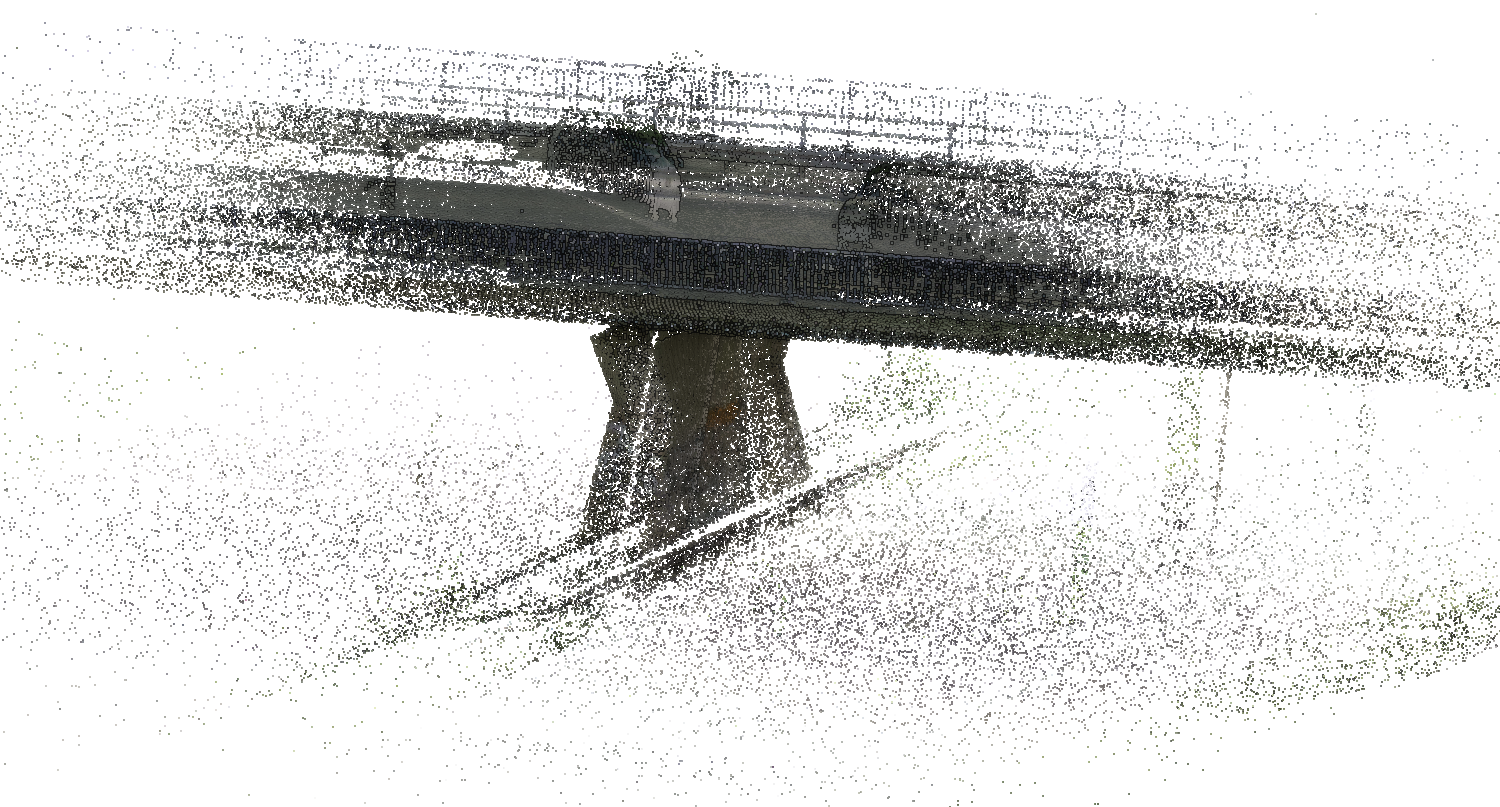}
        \caption{Exponential crop $p_e$}
    \end{subfigure}
    \begin{subfigure}[c]{0.22\textwidth}
        \includegraphics[width=\textwidth]{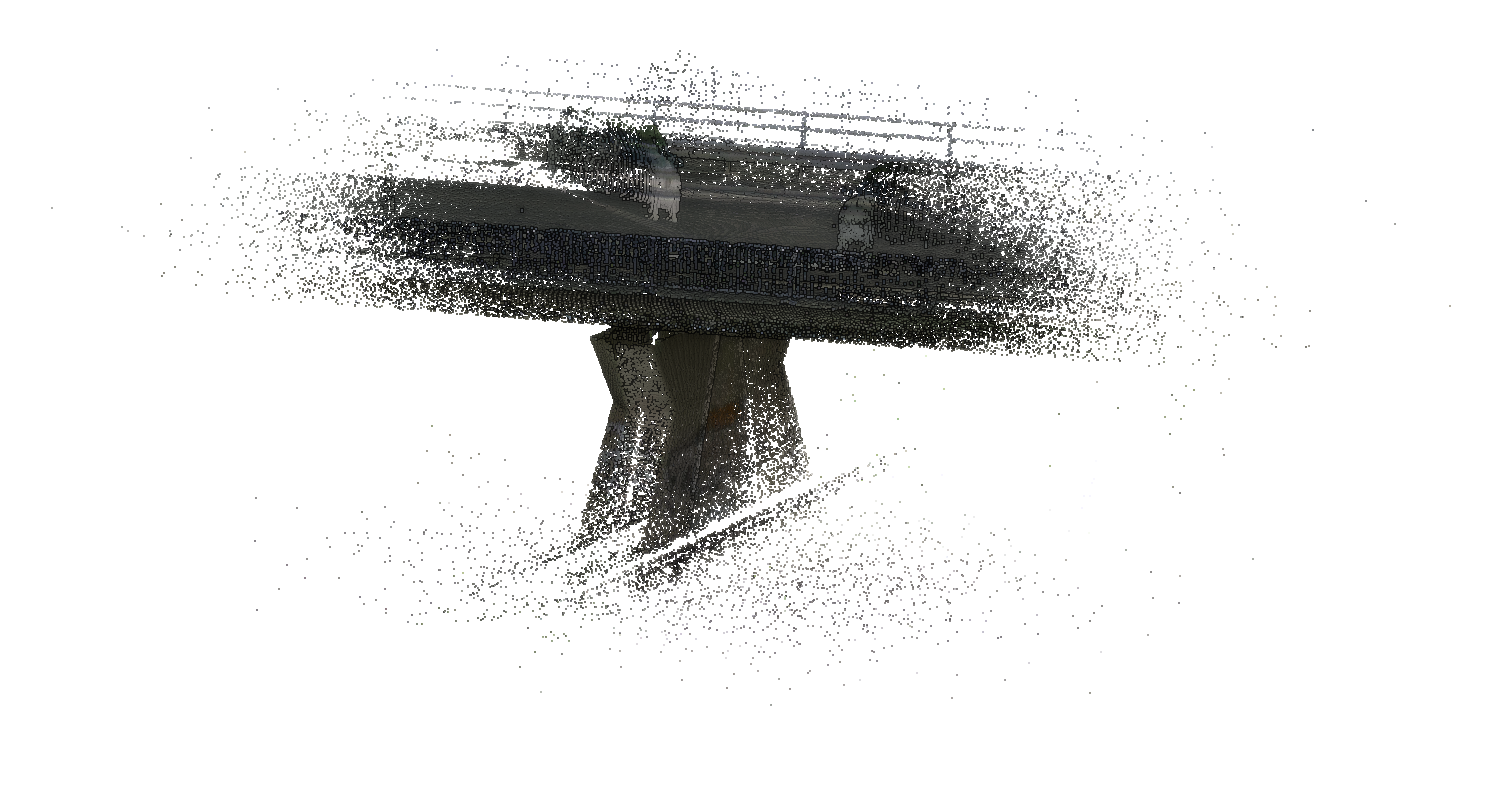}
        \caption{Gaussian crop $p_g$}
    \end{subfigure}
    \begin{subfigure}[c]{0.22\textwidth}
        \includegraphics[width=\textwidth]{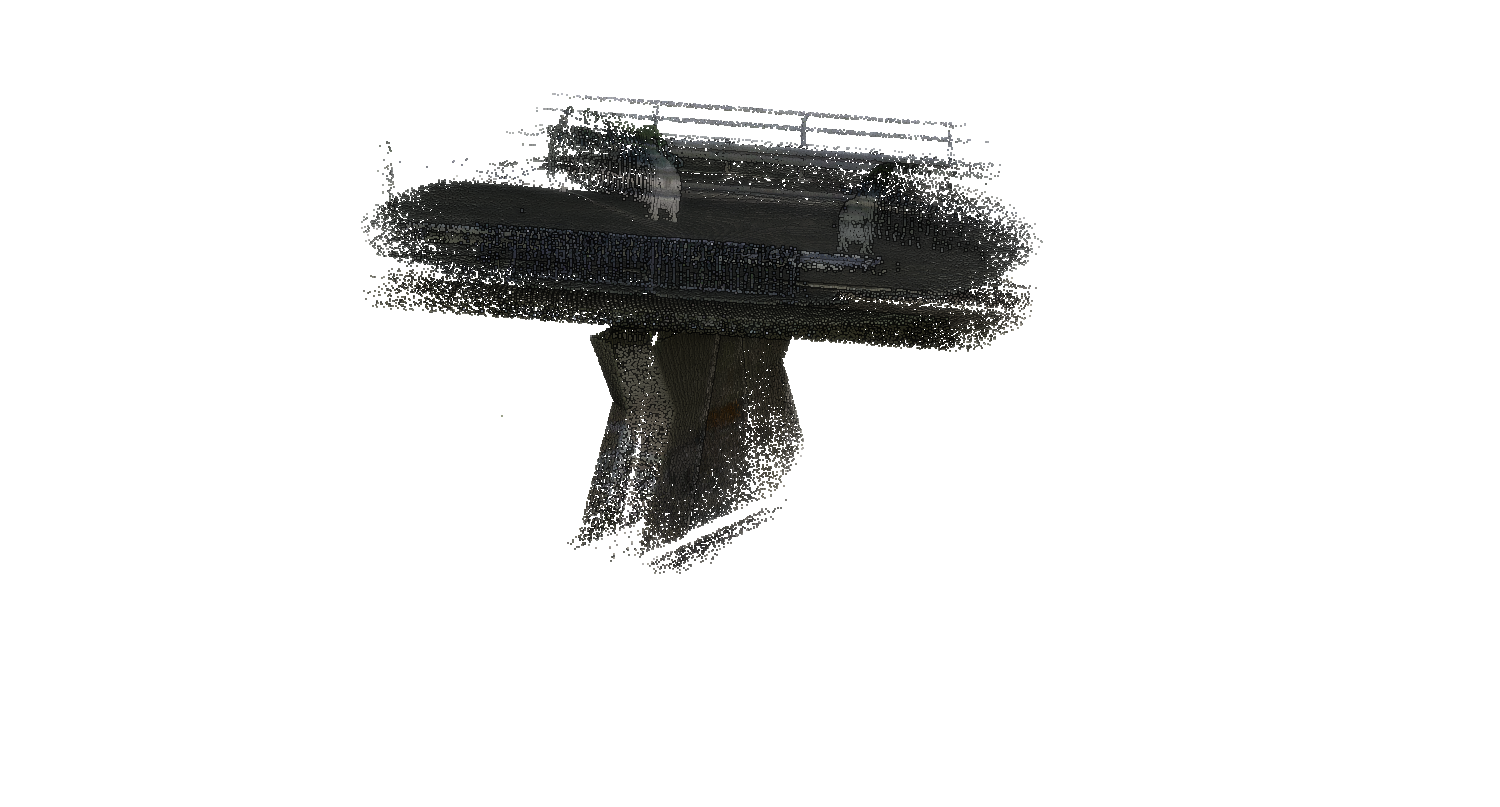}
        \caption{Linear crop $p_l$}
    \end{subfigure}
    \caption{Full point cloud with different subcloud cropping methods. Each crop has approx. 240k points.}
    \label{fig:point_cloud_crops}
\end{figure}

We visualize the probabilities of a point being selected depending on its distance from the crop center within Figure \ref{fig:distributions_over_distance}. It is important to show, that each of the cropping strategies has one critical parameter. This could be either the maximum allowed distance $d_m$, the decay factor $\lambda$ or the scale parameter $\sigma_d$ for the gaussian distribution. To clarify the effects, we visualize each cropping method using a large scene of a bridge from the SemanticBridge dataset in Figure \ref{fig:point_cloud_crops}. All cropping methods use the same center point, which is shown in red in Figure~\ref{fig:point_cloud_crops_a}. To demonstrate the effects of different cropping methods, this example uses cropping parameters that result in each subcloud having approximately 240k points. Although the spherical crop has the most uniform point distribution, it contains the least scene context. The exponential method shows the most context; however, the regions farther away are very sparse, making it difficult to recognize objects. The Gaussian and linear cropping methods seem to strike a balance between showing surrounding context and keeping the point cloud dense enough to recognize objects farther away from the center.

These cropping alternatives were selected for their ability to perform operations with high efficiency. Nevertheless, it's important to recognize that there are other options and more specialized methods besides simple probabilistic cropping.

\subsection{Center point selection}

There are different strategies to define the centers within the full scene in order to split it into subclouds. To ensure segmentation of the full scene, each point must be included in at least one subcloud. One way to accomplish this is to randomly select center points iteratively. The procedure is to initially select a random center. Then, take the points selected according to the cropping strategy and assign the calculated probability to them. The new center is a point with a low assigned probability. This process is repeated until each point has exceeded a threshold probability. It is important to note that, in this approach, dense regions are likely to be selected more often than sparser regions.

As an alternative, the entire point cloud can be divided into an occupancy grid map. The size of the grid depends on the cropping parameter, to ensure full coverage of the scene. If a cell is occupied, the center of the grid cell becomes the center of the subcloud. This procedure can be implemented faster, and it is possible to guarantee the same center points for different cropping strategies, allowing for a fair comparison.

\subsection{3D segmentation}

We evaluate different cropping strategies using three deep learning models for semantic segmentation. One is a discretization-based model, OA-CNNs \cite{peng_oa-cnns_2024}, the second one is the newest version of the Point Transformer model series, PointTransfomer v3 (PTv3) \cite{wu_point_2024} and the third model is the KPConvD \cite{thomas_kpconvx_2024} extension. We chose KPConvD over KPConvX to avoid attention within this architecture and to achieve a more diverse range of architectures. All of these models are state of the art. We do not use pre-trained weights, but instead train all models from scratch.

We configured OA-CNNs using the following parameters: The encoder layers have a depth of $[3, 3, 9, 8]$, whereas the decoder is kept simple, as proposed in the original work. The input features are encoded into 64 features within the embedding stage. The encoder layers use $[62, 62, 128, 256]$ and the decoder ones use $[256, 128, 64, 64]$ features. The downsampling and upsampling modules are implemented using sparse convolution \cite{spconv2022}. Each encoder block comprises a downsampling module, using a stride of 2, and an adaptive aggregator component, followed by two sparse convolutions with a kernel size of 3. Unlike the original work, we do not use depthwise convolution. The adaptive aggregator uses multiple grid sizes and learns how to weight them in order to achieve an adaptive receptive field. PyTorch Geometric \cite{fey_fast_2019} is used to do this efficiently. The point grid sizes for each encoding layer are set to $[16, 32, 64]$ , $[8, 16, 24]$, $[4, 8, 12]$ and $[2, 4, 6]$, starting from the first layer. The decoder layer simply uses upsampling and a single linear layer to align the features, following the original work.

PTv3 is configured using an initial embedding layer with 32 features. The following encoder has four layers each with a depth of $[2,2,6,2]$ with $[64, 128, 256, 512]$ features and $[4,8,16,32]$ heads. The patch size is set to 1024 for all layers of the encoder as well as the decoder. To process this large patch size, memory-efficient flash attention \cite{dao_flashattention_2022, dao_flashattention-2_2023} is used. The decoder layers use a depth of $[2,2,2,2]$ with $[256, 128, 64, 64]$ features and $[16,8,4,4]$ heads. Instead of batch normalization \cite{ioffe_batch_2015}, as used in the original work, we follow \cite{wu_sonata_2025} and replace them with layer normalization \cite{ba_layer_2016}.

A KPConvD block is defined as a linear layer with batch normalization and ReLU activation wrapped around a depthwise KPConv block with a skip connection. The entire model comprises an embedding block with 64 features and 12 neighboring points. The following encoder has a depth of $[2, 2, 12, 4]$ with $[96, 128, 256, 384]$ features and $[16, 20, 20, 20]$ neighbors in each layer. The grid size doubles with each layer. The radius is scaled by a factor of 2.1, and a two-shell kernel disposition of $[1, 14, 28]$ is used. These values follow the original work accordingly. The decoder is kept simple and follows the same idea as described in OA-CNNs \cite{peng_oa-cnns_2024} using only upsampling and single linear layers. The final architecture is similar to the original KPConvD-S one, but with a smaller decoder.

\subsection{Fuse predictions}

Since the scene is divided, predictions from all subclouds must be merged back into the original scene. The most naive approach would be to simply take the predicted point at the time of its final appearance within a subcloud. We follow a similar approach to that proposed in \cite{thomas_kpconv_2019}. First, we use center selection based on the occupancy grid to select subcloud centers regularly. The grid size is defined so that each point is tested multiple times in different subcloud positions. Second, as in \cite{thomas_kpconv_2019}, the predicted probabilities are transformed back into the original scene and averaged.

\section{Experiments}\label{ch:experiments}
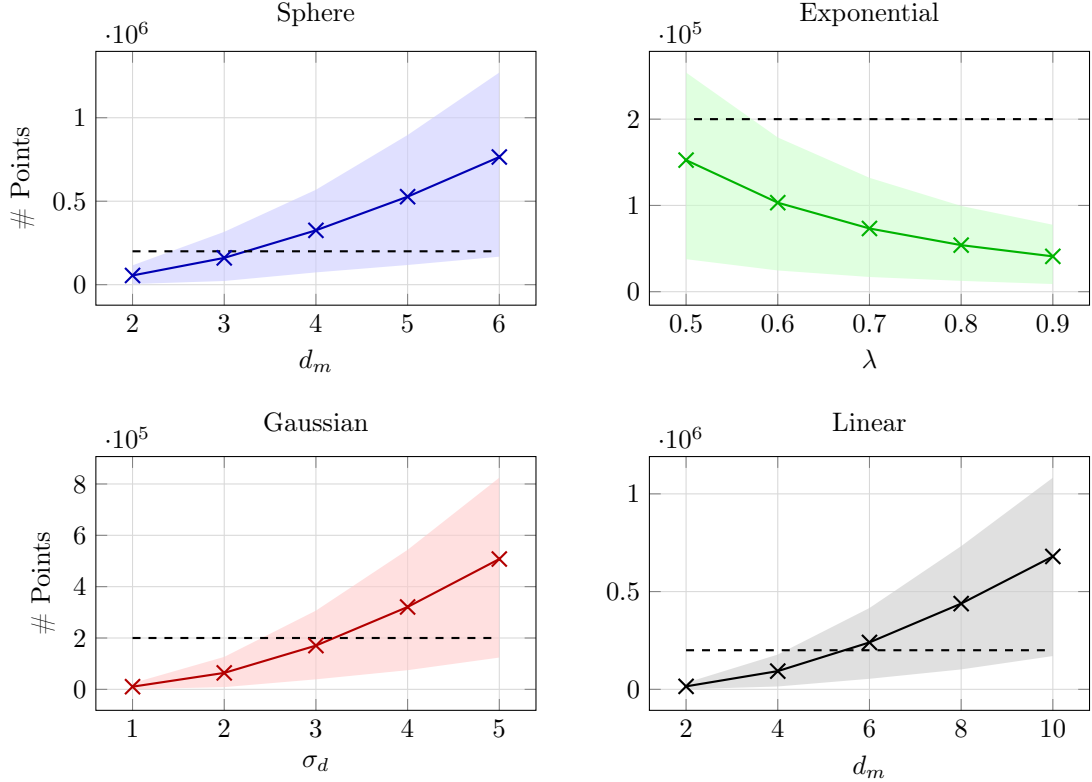
\begin{figure*}[ht]
\centering
    \begin{tikzpicture}
        \begin{groupplot}[
            group style={
                group size=2 by 2,
                horizontal sep=1.5cm,
                vertical sep=2cm
            },
            width=0.45\textwidth,
            height=0.3\textwidth,
            grid=major,
            grid style={gray!30},
            every axis plot/.append style={line width=1.5pt}
        ]
        
        \nextgroupplot[
            title=Sphere,
            ylabel=\# Points,
            xlabel=$d_m$
        ]
        \addplot[name path=sphere_upper, draw=none] coordinates {
            (2, 115790) (3, 316502) (4, 568606) (5, 896410) (6, 1270772)
        };
        \addplot[name path=sphere_lower, draw=none] coordinates {
            (2, 3941) (3, 22203) (4, 73673) (5, 117893) (6, 167697)
        };
        \addplot[blue!20, opacity=0.6] fill between[of=sphere_upper and sphere_lower];
        \addplot[blue!70!black, mark=x, thick, mark size=4pt] coordinates {
            (2, 55233) (3, 160521) (4, 325486) (5, 527117) (6, 764774)
        };
        \addplot[black, dashed, thick] coordinates {(2, 200000) (6, 200000)};
        
        \nextgroupplot[
            title=Exponential,
            xlabel=$\lambda$
        ]
        \addplot[name path=exp_upper, draw=none] coordinates {
            (0.9, 77584) (0.8, 99404) (0.7, 132028) (0.6, 178781) (0.5, 254024)
        };
        \addplot[name path=exp_lower, draw=none] coordinates {
            (0.9, 8893) (0.8, 12554) (0.7, 17054) (0.6, 24606) (0.5, 37691)
        };
        \addplot[green!20, opacity=0.6] fill between[of=exp_upper and exp_lower];
        \addplot[green!70!black, mark=x, thick, mark size=4pt] coordinates {
            (0.9, 40956) (0.8, 53978) (0.7, 73286) (0.6, 103182) (0.5, 152519)
        };
        \addplot[black, dashed, thick] coordinates {(0.9, 200000) (0.5, 200000)};
                
        \nextgroupplot[
            title=Gaussian,
            ylabel=\# Points,
            xlabel=$\sigma_d$
        ]
        \addplot[name path=gauss_upper, draw=none] coordinates {
            (1, 24071) (2, 126388) (3, 306645) (4, 543476) (5, 824429)
        };
        \addplot[name path=gauss_lower, draw=none] coordinates {
            (1, 599) (2, 8626) (3, 38966) (4, 74618) (5, 123851)
        };
        \addplot[red!20, opacity=0.6] fill between[of=gauss_upper and gauss_lower];
        \addplot[red!70!black, mark=x, thick, mark size=4pt] coordinates {
            (1, 10327) (2, 64037) (3, 170392) (4, 320855) (5, 507677)
        };
        \addplot[black, dashed, thick] coordinates {(1, 200000) (5, 200000)};
        
        \nextgroupplot[
            title=Linear,
            xlabel=$d_m$
        ]
        \addplot[name path=lin_upper, draw=none] coordinates {
            (2, 33354) (4, 177912) (6, 416034) (8, 732458) (10, 1082634)
        };
        \addplot[name path=lin_lower, draw=none] coordinates {
            (2, 633) (4, 14551) (6, 53910) (8, 101846) (10, 170303)
        };
        \addplot[black!20, opacity=0.6] fill between[of=lin_upper and lin_lower];
        \addplot[black, mark=x, thick, mark size=4pt] coordinates {
            (2, 15154) (4, 93175) (6, 239934) (8, 438552) (10, 679793)
        };
        \addplot[black, dashed, thick] coordinates {(2, 200000) (10, 200000)};
        
        \end{groupplot}
    \end{tikzpicture}
    
    \caption{Point cardinality for subclouds using a voxel size of 2~cm on the S3DIS dataset. Shaded regions represent the min-max range. Dashed line indicates 200k point line.}
    \label{fig:subcloud_analysis}
\end{figure*}

\subsection{Datasets}

For our approach, we only consider datasets that span a large region and require the scene to be divided into subclouds. Additionally, we aim to validate the method against indoor and outdoor scenes. For this purpose, we use the SemanticBridge \cite{kellner_semanticbridge_2025}, the S3DIS \cite{armeni_3d_2016}, the Paris-Lille-3D \cite{roynard_paris-lille-3d_2018}, and the Toronto3D \cite{tan_toronto-3d_2020} datasets.

The SemanticBridge dataset is specifically designed for segmenting bridge parts. To this end, it contains uncommon classes that are not covered by other datasets. A total of 20 bridges were scanned using a terrestrial laser scanner and annotated using nine classes. The S3DIS dataset includes colored 3D point clouds of six large indoor spaces spanning 6,020 square meters across three buildings. The dataset is also available split into 271 rooms. However, we only use the full scenes to test different cropping methods. The points are densely sampled on the mesh surfaces and annotated with 13 semantic categories. The Paris-Lille-3D dataset was collected using a mobile laser scanner. It contains four outdoor scenes from two cities, spanning approximately 2 kilometers. There is no color information available, only the intensity of the LiDAR. There are many classes within the dataset; however, only nine classes are used for learning and validation purposes. The official benchmark is not suitable for parameter evaluation and is limited to one submission per day. For this reason, we used the Lille2 section for evaluation and submitted only the final result to the benchmark. The Toronto3D dataset was collected using a mobile laser scanner. It contains four outdoor urban scenes, with a total trajectory of 1 km, which are annotated with eight semantic classes. The L002 section is defined as the test split. It contains color and intensity information and is divided into two benchmarks. One uses color information, and the other uses intensity. We only used the intensity benchmark.

We will only use the xyz coordinates as input for the SemanticBridge and the Paris-Lille-3D dataset. For S3DIS, we additionally use color information and for Toronto3D the intensity. We do not use commonly prior calculated features such as normals. This way we only focus on the most general given information. Additionally, we do not use test time augmentation, even though it has been proven to improve performance by more than 2~\% mIoU in \cite{chen_largekernel3d_2023}. Regardless of the dataset, there are various ways to evaluate performance. This can be done either on a subcloud basis or after the subclouds have been fused to create a full cloud. To make it easier to compare the different strategies, we evaluate them on a subcloud basis. To make a comparison with the final benchmarks, however, we take the fusion of the subclouds into account and evaluate the full cloud.

\subsection{Implementation details}

Our implementation is mainly based on NumPy \cite{harris_array_2020} and PyTorch \cite{paszke_pytorch_2019}. To accelerate the cropping methods we make use of numba \cite{lam_numba_2015}. To train our segmentation models, we used AdamW \cite{loshchilov_decoupled_2017} as an optimizer with an initial learning rate of 0.0005, and a one-cycle scheduler \cite{smith_super-convergence_2017} with a maximum learning rate of 0.005 and five warm up steps. For PTv3, the attention blocks are trained with a learning rate reduced by a factor of 10. The models are trained for a total of 100~epochs. We adjust the batch size and accumulate multiple forward passes before propagating the gradient backward. This allows us to use a batch size of 16 while reducing the GPU memory usage. The training was conducted on a single NVIDIA A100 GPU. We solely use the cross-entropy as the loss function with label smoothing \cite{szegedy_rethinking_2015}. We use a smoothing factor of 0.05.

During training, each subcloud is augmented. Up to 20~\% of the points are randomly dropped. If color information is used, up to 30~\% of the colors are dropped and Gaussian noise with a standard deviation of $\sigma=0.01$ is added after normalization. The point coordinates are scaled by a random factor between 0.9 and 1.1, rotated around the z-axis, flipped against the x- and y-axes, and shifted along each axis. Finally, Gaussian noise with $\sigma = 0.005$ is added. Each augmentation, except for dropping points and color, is applied with a 50~\% probability.

\subsection{Ablation studies and analysis}

We conducted a prior hyperparameter analysis to investigate the impact of parameters such as the initial input voxel size and the cropping parameters, as well as the robustness of the trained models in the face of changes to these parameters. Using the OA-CNNs model, the S3DIS dataset for indoor scenes, and the SemanticBridge dataset for outdoor scenes, we tune hyperparameters for the best subcloud semantic segmentation performance.

\textbf{Point cardinality.} To illustrate the issue of how many points are acquired for one forward pass, we used the S3DIS dataset to calculate subclouds for different cropping strategies and parameters. The results are shown in Figure \ref{fig:subcloud_analysis}. It should be noted that the same center points are used for all methods, and the full point cloud is downsampled using a grid size of 2~cm. It is clear that the amount of points in the spherical approach increases drastically with an increasing input radius. On average, there are $\sim200k$~points within a subcloud with an input radius of 3~m. The same number of points within a subcloud can be achieved by using a $\lambda = 0.46$ for exponential, a variance of $\sigma = 3.2$ for Gaussian, and a $d_m=5.6$ for linear cropping.

However, in order to examine the spatial size of the crop, the maximum distance between the selected center and all other points within the subcloud must be calculated. These measurements show how far the subcloud extends and how much surrounding context is included. This can easily be done by $d_\text{max} = \text{max}(\|\mathbf{x}_i - \mathbf{x}_c\|), \quad i = 1, 2, \ldots, N$. For spherical and linear systems, $d_{max}$ is equivalent to the selected cropping parameter. Exponential and Gaussian cropping achieve much larger extents of $d_\text{max}$, reaching approximately 30 and 12 meters, respectively.

\textbf{Voxel size.} To investigate into the impact of the voxel size, we only use the spherical subcloud calculation and train the model on different voxel sizes. As the Table \ref{tab:performance_over_voxel_sizes} clearly shows, the initial voxel size has a significant impact. The performance is negatively impacted by overly focusing on minor details, which is exacerbated by using overly small sizes. However, the network is unable to distinguish objects due to the smooth out details being too large.

\begin{table}
    \centering
    \footnotesize
    \caption{Influence on training using different voxel sizes and same radius. S3DIS uses a radius of 4~m and SemanticBridge of 8~m. Evaluated on subcloud basis using mIoU.}
    \label{tab:performance_over_voxel_sizes}
    \begin{tabular}{@{} c r r @{}}
        \hline
        Voxel size & S3DIS & SemanticBridge \\
        \hline
        0.01 & 58.1 & --- \\
        0.02 & \textBF{64.0} & --- \\
        0.03 & 62.9 & --- \\
        0.04 & 61.1 & --- \\
        0.05 & 60.0 & 65.4 \\
        0.06 & 57.9 & --- \\
        0.08 & --- & 67.4 \\
        0.10 & --- & \textBF{68.6} \\
        0.12 & --- & 66.8 \\
        0.15 & --- & 65.7 \\
        \hline
    \end{tabular}
\end{table}

\begin{figure}
    \centering
    \begin{tikzpicture}
        \begin{groupplot}[
            group style={
                group size=2 by 1,
                horizontal sep=0.5cm,
                x descriptions at=edge bottom
            },
            width=4.8cm, height=4.8cm,
            grid = major,
            grid style={line width=0.3pt, draw=gray!40, dashed},
            axis lines=left,
            ymin = 55, ymax = 82,
            xmin=0,
            tick label style={font=\footnotesize},
            ticklabel shift={-3pt},
            clip=false, 
            legend pos=south east,
            legend style={
                at={(0.5,-0.3)},
                anchor=north,
                legend columns=1,
                font=\scriptsize,
            },
        ]

        \nextgroupplot[
            xlabel={$r$}, 
            xtick={1,2,3,4,5,6,7,8,9},
        ]
            \node[anchor=south west, inner sep=2pt] at (rel axis cs:-0.12,1) {mIoU};
            \addplot[color=blue!70!black, line width=1pt, mark=*] coordinates {
                 (1,57.39) (2,62.8) (3, 63.67) (4, 63.73) (5, 63.96) (6, 63.14)
            };
            \addlegendentry{S3DIS $p_s$}
            
            \addplot[color=cyan!70!black, line width=1pt, mark=diamond*] coordinates {
                 (3, 65.7) (4, 68.4) (5, 69.5) (6, 69.7) (7, 71.9) (8, 72.3) (9, 71.2)
            };
            \addlegendentry{SemanticBridge $p_s$}

        \nextgroupplot[
            xlabel={$\sigma$}, 
            xtick={1,2,3,4,5},
            ylabel={}, 
        ]
            \addplot[color=red!70!black, line width=1pt, mark=*] coordinates {
                 (1,65.48) (1.5, 66.5) (2,64.01) (3, 61.24)
            };
            \addlegendentry{S3DIS $p_g$}

            \addplot[color=orange!80!black, line width=1pt, mark=diamond*] coordinates {
                 (2, 72.2) (2.5, 74.3) (3, 77.8) (3.5, 80.1) (4, 79.7) (4.5, 78.9) (5, 78)
            };
            \addlegendentry{SemanticBridge $p_g$}

        \end{groupplot}
    \end{tikzpicture}
    \caption{Influence on training using different point cropping strategies ($p_s$ and $p_g$) and parameters (m). Evaluated on subcloud basis using mIoU.}
    \label{fig:variing_input_params}
\end{figure}
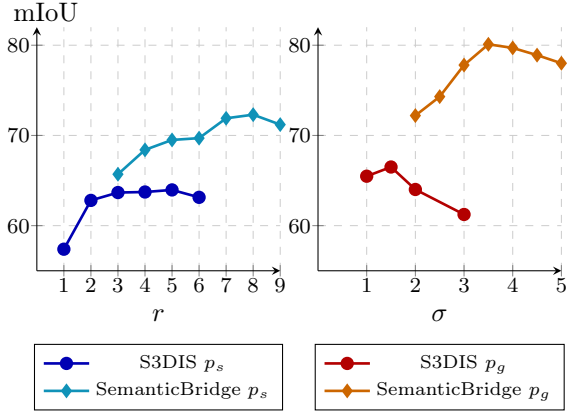

\textbf{Cropping parameter.} To investigate the influence of the cropping parameters, we trained the model multiple times using different settings. For S3DIS, we use an input radius between 1 to 6 m for spherical cropping, and for SemanticBridge, we use an input radius between 3 to 9 m. For Gaussian cropping, we use $\sigma$ values between 1 to 3 for S3DIS and 2 to 4 for SemanticBridge with 0.5 as step size. The results can be seen in Figure \ref{fig:variing_input_params}. The figure shows that increasing the outer bound of the cropping method results in an increase in performance. Eventually, however, performance saturates and decreases. This could happen either because further context-relevant information is not obtained or because of the internal receptive field of the model itself.

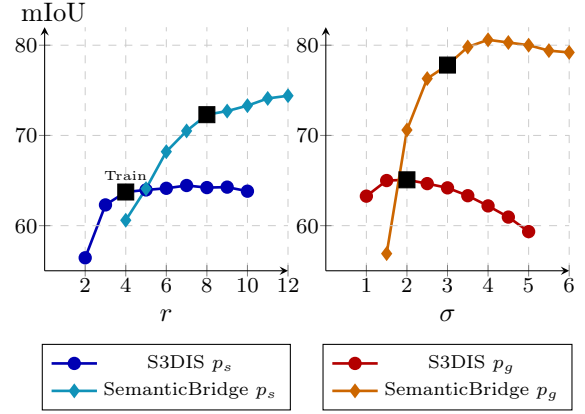
\begin{figure}
    \centering
    \begin{tikzpicture}
        \begin{groupplot}[
            axis on top,
            group style={
                group size=2 by 1,
               horizontal sep=0.5cm,
                x descriptions at=edge bottom
            },
            width=4.8cm, height=4.8cm,
            grid = major,
            grid style={line width=0.3pt, draw=gray!40, dashed},
            axis lines=left,
            ymin = 55, ymax = 82,
            xmin=0,
            tick label style={font=\footnotesize},
            ticklabel shift={-3pt},
            clip=false, 
            legend pos=south east,
            legend style={
                at={(0.5,-0.3)},
                anchor=north,
                legend columns=1,
                font=\scriptsize,
            },
        ]

        \nextgroupplot[xlabel={$r$},
                       xtick={2,4,6,8,10,12},
                       ]
            \node[anchor=south west, inner sep=2pt] at (rel axis cs:-0.12,1) {mIoU};
            \addplot[color=blue!70!black, line width=1pt, mark=*] coordinates {
                (2, 56.44) (3, 62.3) (4, 63.73) (5, 63.96) (6, 64.14) 
                (7, 64.45) (8, 64.22) (9, 64.27) (10, 63.82)
            };
            \addlegendentry{S3DIS $p_s$}
            
            \addplot[color=cyan!70!black, line width=1pt, mark=diamond*] coordinates {
                (4, 60.6) (5, 64.1) (6, 68.2) (7, 70.5) (8, 72.3) (9, 72.7) (10, 73.3) (11, 74.1) (12,74.4)
            };
            \addlegendentry{SemanticBridge $p_s$}

            \addplot[only marks, mark=square*, mark size=3pt, color=black, forget plot] coordinates {(4, 63.73) (8, 72.3)};
            \node[anchor=south] at (axis cs:4, 64) {\tiny Train};

        \nextgroupplot[xlabel={$\sigma$},
                       xtick={1,2,3,4,5,6},
                       ylabel={},
                       ]
            \addplot[color=red!70!black, line width=1pt, mark=*] coordinates {
                (1, 63.27) (1.5, 65) (2, 65.08) (2.5, 64.66) (3, 64.19) (3.5, 63.32) (4, 62.19) (4.5, 60.95) (5, 59.34)
            };
            \addlegendentry{S3DIS $p_g$}

            \addplot[color=orange!80!black, line width=1pt, mark=diamond*] coordinates {
                (1.5, 56.9) (2, 70.6) (2.5, 76.3) (3, 77.8) (3.5, 79.8) (4, 80.6) (4.5, 80.3) (5, 80.0) (5.5, 79.4) (6, 79.2)
            };
            \addlegendentry{SemanticBridge $p_g$}

            \addplot[only marks, mark=square*, mark size=3pt, color=black, forget plot] coordinates {(2, 65.08) (3, 77.8)};
        \end{groupplot}
    \end{tikzpicture}
    \caption{Performance validation using different cropping parameters then trained on. Squares indicate the specific parameters used during training.}
    \label{fig:radius_performance_analysis}
\end{figure}

We evaluate how well the trained model performs when the size of the subcloud changes during inference. To this end, we vary the radius $r$ of the sphere and the $\sigma$ for Gaussian cropping. The models are trained using $r=4$ and $\sigma=2$ for S3DIS. For SemanticBridge we use a radius $r=8$ and a $\sigma=3$. We use the same subcloud center points for all the different input parameters and evaluate them on a subcloud basis. The results can be seen in Figure \ref{fig:radius_performance_analysis}. Interestingly, the evaluation results improve when a higher cropping parameter (r or $\sigma$) is used for evaluation than for training, compared to using the same value for both training and evaluation. This seems to be the case for all tested methods and datasets, except for Gaussian cropping on S3DIS. When using a cropping method other than the spherical one, the distribution of points changes with the cropping parameter. When paired with a small voxel size, the change in distribution is large enough that the model cannot rely on its learned pattern. Nevertheless, it shows that a larger subcloud leads to more surrounding context, which improves performance. This also explains why, during training, only subclouds are used for the S3DIS dataset to save memory, while the full room can be used for inference with improved performance \cite{wu_point_2024, thomas_kpconvx_2024}.

\begin{figure}
    \centering
    \begin{tikzpicture}
        \begin{axis}[
            xlabel = {vs},
            axis on top,
            width=7cm, height=7cm,
            grid = major,
            grid style={line width=0.3pt, draw=gray!40, dashed},
            axis lines=left,
            clip=false, 
            ymin = 15, ymax = 79,
            xmin = 0, xmax= 15,
            tick label style={font=\small},
            label style={font=\normalsize},
            legend pos=south east,
            legend style={
                at={(0.5,-0.2)},
                anchor=north,
                legend columns=2,
                font=\scriptsize,
            },
        ]

        \node[anchor=south west, inner sep=2pt] at (rel axis cs:-0.12,1) {mIoU};

        \addplot[color=blue!70!black, line width=1pt, mark=*] coordinates {
            (1, 25.16) (2, 63.73) (3, 41.05) (4, 26.97) (5, 19.68)
        };
        \addlegendentry{S3DIS $p_s$}
        
        \addplot[color=cyan!70!black, line width=1pt, mark=diamond*] coordinates {
            (3, 46.41) (4, 63.85) (5, 70.06) (6, 67.36) (7, 65.11)  (8, 63.13)
        };
        \addlegendentry{SemanticBridge $p_s$}

        \addplot[only marks, mark=square*, mark size=3pt, color=black, forget plot] coordinates {(2, 63.73) (5, 70.06)};
        \node[anchor=south] at (axis cs:2, 66) {\tiny Train};

        \addplot[color=red!70!black, line width=1pt, mark=*] coordinates {
           (1, 27.36)(2, 65.08) (3, 38.95) (4, 23.46) (5, 16.67)
        };
        \addlegendentry{S3DIS $p_g$}

        \addplot[color=orange!80!black, line width=1pt, mark=diamond*] coordinates {
            (3, 51.48) (4, 70.93) (5, 72.17) (6, 67.49) (7, 64.44) (8, 58.37)
        };
        \addlegendentry{SemanticBridge $p_g$}

        \addplot[color=cyan!70!black, line width=1pt, mark=diamond*] coordinates {
            (7, 59.5) (8, 64.6) (9, 68.4) (10, 72.3) (11, 69.4) (12, 69.3) (13, 69) (14, 68) (15, 64.4)
        };
        \addplot[color=orange!80!black, line width=1pt, mark=diamond*] coordinates {
            (7, 70) (8, 76.6) (9, 77) (10, 77.8) (11, 76) (12, 71.5) (13, 67.6) (14, 62.2) (15, 58.1)
        };

        \addplot[only marks, mark=square*, mark size=3pt, color=black, forget plot] coordinates {(2, 65.08) (5, 72.17) (10, 72.3) (10, 77.8)};
        
        \end{axis}
    \end{tikzpicture}
    \caption{
        Performance analysis of voxel size (cm) variation on S3DIS and SemanticBridge dataset. The models were trained using a sphere radius of 4~m and 8~m. The gaussian cropping used a $\sigma = 2$ and $\sigma = 3$. The voxel size during training was set to 2~cm, 5~cm and 10~cm. Evaluated on subcloud basis.
    }
    \label{fig:voxel_size_performance_analysis}
\end{figure}
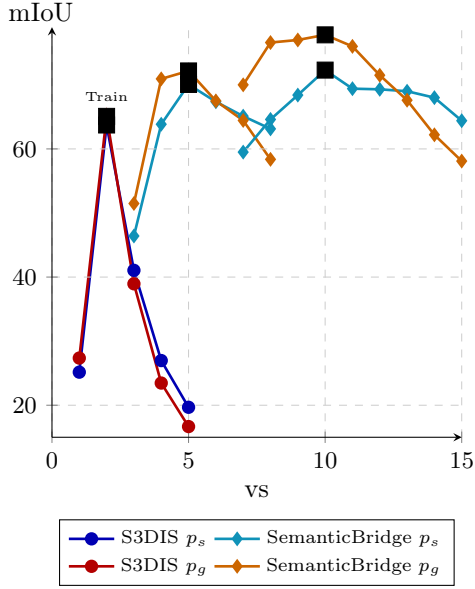

\textbf{Input robustness.} While the model is robust against small changes in the cropping parameter and can benefit from adapting the parameter to increase the outer bound of the subcloud, it is highly sensitive to changes of the voxel size used during training. The results can be seen in Figure \ref{fig:voxel_size_performance_analysis}. This is expected, given that the voxel size defines the grid to which the sparse convolution is applied. Changes within the grid directly affect the pattern, which has a negative effect on the learned filters. The models trained on the Gaussian cropping are more sensitive to changes, as is illustrated. Additionally, the model's sensitivity to small changes is directly proportional to the voxel size utilized during training.

\begin{figure}
    \centering
    \begin{tikzpicture}
        \begin{axis}[
            xlabel = {$\tau_p$},
            axis on top,
            width=7cm, height=7cm,
            grid = major,
            grid style={line width=0.3pt, draw=gray!40, dashed},
            axis lines=left,
            clip=false, 
            xmin = 0, xmax = 1,
            ymin = 55, ymax = 76,
            xtick = {0.2, 0.4, 0.6, 0.8},
            tick label style={font=\small},
            label style={font=\normalsize},
            legend pos=south east,
            legend style={
                at={(0.5,-0.25)},
                anchor=north,
                legend columns=2,
                font=\scriptsize,
            },
        ]
        \node[anchor=south west, inner sep=2pt] at (rel axis cs:-0.12,1) {mIoU};
        
        \addplot[color=blue!70!black, line width=1pt, mark=*,] coordinates {
            (0.1, 59.58) (0.2, 61.69) (0.3, 62.55) (0.4, 63.15) (0.5, 63.48) (0.6, 63.64) (0.7, 63.71) (0.8, 63.75) (0.9, 63.79)
        };
        \addlegendentry{S3DIS $p_s$}

        \addplot[color=cyan!70!black, line width=1pt, mark=diamond*] coordinates {
            (0.1, 61.13) (0.2, 62.57) (0.3, 63.77) (0.4, 64.68) (0.5, 65.2) (0.6, 65.52) (0.7, 65.79) (0.8, 65.86)  (0.9, 65.87)
        };
        \addlegendentry{SemanticBridge $p_s$}

        \addplot[color=red!70!black, line width=1pt, mark=*] coordinates {
            (0.1, 63.68) (0.2, 66.23) (0.3, 66.98) (0.4, 67.34) (0.5, 67.39) (0.6, 67.34) (0.7, 67.32) (0.8, 67.52) (0.9, 67.81)
        };
        \addlegendentry{S3DIS $p_g$}


        \addplot[color=orange!80!black, line width=1pt, mark=diamond*] coordinates {
            (0.1, 63.98) (0.2, 67.6) (0.3, 69.36) (0.4, 70.53) (0.5, 71.12) (0.6, 71.74) (0.7, 72.3) (0.8, 72.61)  (0.9, 72.77)
        };
        \addlegendentry{SemanticBridge $p_g$}

        \end{axis}
    \end{tikzpicture}
    \caption{The influence of the probability between all points in the subcloud and the center point $\mathbf{x}_c$ on the distribution of the segmentation error. Evaluated on subcloud basis.}
    \label{fig:distance_influence_in_sub_cloud}
\end{figure}
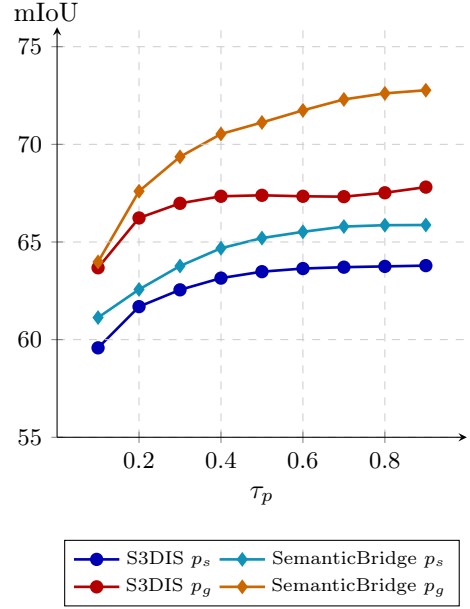

\textbf{Distance sensitivity.} The center of each subcloud is determined by the center point $\mathbf{x}_c$ selected from the full cloud. This means that the subcloud provides the most context in the center and raises the question of whether the model performs better close to the selected center than farther away. For Gaussian cropping, the probability of a point being selected is already calculated, and for spherical cropping, it can easily be calculated using the distance $d$ and the specified maximum distance $d_m$ using $p = 1 - \frac{d}{d_m}$. The metric is computed based on the selection probability of each point, where the subset of points considered is constrained by a probability threshold $\tau_p$ such that $p \leq \tau_p$. The parameter space for this threshold is defined as $\mathcal{T}_p = \{ \tau_p \in \mathbb{R} \mid \tau_p = 0.1n, n \in \{2, 3, \dots, 9\} \}$. The results are visualized in Figure~\ref{fig:distance_influence_in_sub_cloud}. It indicates that distance has an effect on performance, and proves that points close to the center are more likely to be predicted correctly than those further away.

\textbf{Cross cropping.} After training the model using only one cropping method, we evaluate the trained model using all other cropping methods to see if the cropping and the given distribution affect the learned patterns. The results are shown in Table \ref{tab:cross_sampling}. Interestingly, spherical cropping is the only method where the performance is significantly affected by changing the cropping strategy during evaluation. This makes sense considering the data. While spherical cropping aims to keep the points as uniformly distributed as possible, the other methods degrade this distribution. It shows that OA-CNNs can learn non-uniform distributions, and that they are less affected by changes once learned.

\begin{table}[ht!]
    \centering
    \footnotesize
    \caption{Cross cropping evaluation. Evaluated on subcloud basis on S3DIS Area5 using mIoU.}
    \label{tab:cross_sampling}
    \begin{tabular}{l|cccc}
    \hline
    \multicolumn{5}{c}{S3DIS} \\
    \hline
    \multicolumn{1}{c|}{\diagbox{Train}{Val}} & $p_s$ & $p_e$ & $p_g$ & $p_l$ \\
    \hline
    $p_s$ & \textbf{64.1} & 39.4 & 53.1 &  60.1 \\
    $p_e$ & 61.1 & 62.1 & 64.5 & \textbf{64.8} \\
    $p_g$ & 66.5 & 65.9 & 67.4 & \textbf{68.3} \\
    $p_l$ & 67.4 & 64.4 & 67.8 & \textbf{68.4} \\
    \hline
    \multicolumn{5}{c}{SemanticBridge} \\
    \hline
    $p_s$ & \textbf{72.3} & 34.0 & 45.3 & 52.1 \\
    $p_e$ & 67.1 & \textbf{74.6} &  72.1 & 73.5 \\
    $p_g$ & 77.4 & 75.1 & 77.8 & \textbf{79.8} \\
    $p_l$ & \textbf{75.4} & 57.6 & 69.6 & 75.3 \\
    \hline
    \end{tabular}
\end{table}

\subsection{Results}\label{ch:results}
Initial evaluations have concentrated on the subclouds themselves and the identification of suitable parameters. Subsequently, the subclouds will be combined to the original full cloud to evaluate the performance. In addition, the Paris-Lille-3D dataset and the PTv3 model will be incorporated. In the previous experiments, it was observed that exponential cropping introduced excessive sparsity, particularly in outdoor scenes. For this reason, this cropping is not used in the following experiments and the focus is only on the other three strategies. Adapting the cropping parameter to increase the spatial size of the subcloud after training the model has been shown to improve performance on the crop itself. However, we note that this was not done in subsequent experiments. In all experiments, the spherical cropping method serves as our baseline for comparing alternative cropping methods. The values reported in this section are from the model that produced the best results out of three trials. This is also why the values may differ slightly from those of previous experiments.

Based on the findings of the previous analysis, we selected the following parameter setup. The number of points in the crop is similar, except for the Gaussian one, where approximately half of the points are on average subclouds. For the S3DIS dataset, we use $d_m = 3$ for spherical cropping, $d_m = 6$ for linear cropping, and $\sigma_d = 1.7$ for Gaussian cropping. The initial voxel size is 2~cm. We use the same setup for all other datasets. For cropping, we use $d_m = 8$ for spherical, $d_m = 12$ for linear, and $\sigma_d = 3.5$ for Gaussian. The initial voxel size is 10~cm. It is important to note that we did not optimize these parameters to achieve the best performance, but rather to enable comparison of the cropping approaches.

The results for SemanticBridge dataset are shown in Table \ref{tab:semantic_bridge_results}. For this dataset, alternative cropping methods with a larger outer bound improve the performance of all models. This is because of the classes given. For example, distinguishing between an abutment and a pillar depends heavily on the surrounding context. The validity of this claim can be confirmed by examining the class IoUs. In every instance, there is an increase. Furthermore, merging the subclouds into the original scene improves the final performance in all cases, regardless of the model or cropping method used. This is true not only within this dataset, but also within all the other datasets used. This is due to the multiple predictions of points that are part of multiple subclouds. We have seen that misclassified points tend to be close to the outer bound. However, smartly choosing the centers for cropping creates a certain overlap, which reduces this effect. Compared to other methods within this dataset the gaussian cropping proposes a new SOTA with more than 80~\% mIoU.

\begin{table*}[]
    \centering
    \footnotesize
    \setlength{\tabcolsep}{4pt}
    \caption{Results on SemanticBridge test dataset. Evaluation is done on subclouds (S) and on the final full cloud (F).}
    \label{tab:semantic_bridge_results}
\begin{tabular}{ccccccccccccc}
\toprule
    \rotatebox{70}{Model} & \rotatebox{70}{Cropping} & \rotatebox{70}{Evaluation} & \rotatebox{70}{Unlabeled} & \rotatebox{70}{Underground} & \rotatebox{70}{High Veg.} & \rotatebox{70}{Abutment} & \rotatebox{70}{Superstructure} & \rotatebox{70}{Deck} & \rotatebox{70}{Railing} & \rotatebox{70}{Traffic Sign} & \rotatebox{70}{Pillar} & \rotatebox{70}{\textbf{mIoU}} \\
    \midrule
    OA-CNNs & $p_s$ & S & 83.2 & 76.5 & 87.2 & 54.2 & 92.0 & \textbf{89.6} & \textbf{76.9} & 36.0 & 61.1 & 72.3 \\
     & $p_g$ & S & \textbf{84.6} & 78.1 & 86.6 & \textbf{67.0} & 91.7 & 87.2 & 75.9 & \textbf{53.9} & 75.0 & \textbf{77.8} \\
     & $p_l$ & S & 82.9 & \textbf{78.4} & \textbf{87.7} & 62.6 & \textbf{92.5} & 89.5 & 76.8 & 28.1 & \textbf{77.0} & 75.1 \\
    \hdashline
    PTv3 & $p_s$ & S & \textbf{82.8} & 77.5 & 86.8 & 49.8 & \textbf{91.6} & \textbf{89.7} & 74.8 & 38.9 & 49.5 & 71.3 \\
     & $p_g$ & S & 78.2 & 79.4 & 87.2 & \textbf{58.7} & 88.7 & 85.9 & 70.8 & \textbf{41.4} & 56.9 & 71.9 \\
     & $p_l$ & S & 81.6 & \textbf{79.7} & \textbf{88.3} & 55.3 & 91.3 & 88.7 & \textbf{75.4} & 40.9 & \textbf{62.4} & \textbf{73.7} \\
    \hdashline
    KPConvD & $p_s$ & S & \textbf{91.2} & 77.5 & 88.2 & 48.7 & 88.3 & \textbf{90.1} & \textbf{81.5} & \textbf{47.3} & 52.5 & 73.9 \\
     & $p_g$ & S & 89.2 & 78.1 & 87.3 & \textbf{77.5} & 89.0 & 87.3 & 75.6 & 38.1 & \textbf{73.8} & \textbf{77.3} \\
     & $p_l$ & S & 90.6 & \textbf{79.6} & \textbf{88.7} & 56.9 & \textbf{90.6} & 89.5 & 79.3 & 46.0 & 61.7 & 75.9 \\
    \midrule
    OA-CNNs & $p_s$ & F & 87.2 & 78.4 & 85.9 & 50.1 & 93.8 & \textbf{90.9} & \textbf{79.7} & 59.3 & 59.4 & 76.1 \\
     & $p_g$ & F & 87.0 & \textbf{80.5} & 86.4 & \textbf{69.7} & 93.3 & 88.2 & 79.5 & \textbf{60.2} & 76.7 & \textbf{80.2} \\
     & $p_l$ & F & \textbf{88.1} & 79.8 & \textbf{86.6} & 63.0 & \textbf{93.9} & 90.7 & \textbf{79.7} & 25.6 & \textbf{80.1} & 76.4 \\
    \hdashline
    PTv3 & $p_s$ & F & \textbf{88.8} & 80.2 & 85.7 & 50.5 & 93.0 & \textbf{90.8} & 76.6 & 45.5 & 46.9 & 73.1 \\
     & $p_g$ & F & 85.9 & \textbf{81.6} & 86.8 & \textbf{63.0} & 90.7 & 88.2 & 76.6 & \textbf{58.1} & 57.6 & \textbf{76.5} \\
     & $p_l$ & F & 88.3 & 80.9 & \textbf{87.4} & 53.0 & \textbf{93.2} & 89.8 & \textbf{78.5} & 52.5 & \textbf{61.0} & 76.1 \\
    \hdashline
    KPConvD & $p_s$ & F & \textbf{94.2} & 79.2 & 86.5 & 47.6 & 90.5 & \textbf{90.6} & \textbf{82.0} & \textbf{61.8} & 50.1 & 75.8 \\
     & $p_g$ & F & 90.4 & 80.2 & 86.4 & \textbf{82.5} & 90.5 & 88.0 & 77.7 & 53.4 & \textbf{74.1} & \textbf{80.4} \\
     & $p_l$ & F & 91.5 & \textbf{80.6} & \textbf{87.4} & 53.9 & \textbf{91.8} & 89.5 & 80.3 & 61.5 & 59.3 & 77.4 \\
    \bottomrule
\end{tabular}
\end{table*}

\begin{table*}[]
    \centering
    \footnotesize
    \setlength{\tabcolsep}{4pt}
    \caption{Results on S3DIS dataset (entire Area5). Evaluation is done on subclouds (S) and on the final full cloud (F).}
    \label{tab:s3dis_results}
\begin{tabular}{ccccccccccccccccc}
    \toprule
    \rotatebox{70}{Model} & \rotatebox{70}{Cropping} & \rotatebox{70}{Evaluation} & \rotatebox{70}{Ceiling} & \rotatebox{70}{Floor} & \rotatebox{70}{Wall} & \rotatebox{70}{Beam} & \rotatebox{70}{Column} & \rotatebox{70}{Window} & \rotatebox{70}{Door} & \rotatebox{70}{Table} & \rotatebox{70}{Chair}& \rotatebox{70}{Sofa} & \rotatebox{70}{Bookcase} & \rotatebox{70}{Board} & \rotatebox{70}{Clutter} & \rotatebox{70}{\textbf{mIoU}} \\
    \midrule
    OA-CNNs & $p_s$ & S & 92.7 & 97.1 & 83.0 & 0.0 & 29.3 & 45.2 & 77.8 & 74.9 & 89.1 & 61.6 & \textbf{76.8} & 63.8 & \textbf{62.1} & 65.7 \\
     & $p_g$ & S & \textbf{93.3} & \textbf{97.7} & 83.4 & 0.0 & \textbf{37.3} & 53.2 & 80.7 & \textbf{79.2} & 89.7 & 71.8 & 74.6 & 63.7 & 57.6 & 67.9 \\
     & $p_l$ & S & 92.7 & 97.3 & \textbf{85.0} & 0.0 & 32.4 & \textbf{53.7} & \textbf{82.8} & 78.6 & \textbf{91.7} & \textbf{74.0} & 75.1 & \textbf{68.2} & 61.8 & \textbf{68.7} \\
    \hdashline
    PTv3 & $p_s$ & S & \textbf{94.9} & \textbf{97.9} & \textbf{85.2} & 0.0 & \textbf{45.3} & \textbf{55.8} & 72.8 & \textbf{77.5} & 87.8 & 62.5 & \textbf{78.8} & \textbf{63.7} & \textbf{61.6} & \textbf{67.9} \\    
     & $p_g$ & S & 92.2 & \textbf{97.9} & 81.6 & 0.0 & 31.9 & 46.4 & \textbf{76.9} & 71.8 & \textbf{89.2} & \textbf{70.9} & 73.3 & 48.4 & 52.1 & 64.1 \\    
     & $p_l$ & S & 92.6 & 96.7 & 84.6 & 0.0 & 32.6 & 40.2 & 74.8 & 72.1 & \textbf{89.2} & 60.5 & 72.6 & 42.2 & 55.0 & 62.5 \\
    \hdashline
    KPConvD & $p_s$ & S & \textbf{94.0} & 96.5 & 84.4 & 0.0 & 34.1 & \textbf{58.7} & 68.1 & 79.6 & 90.0 & 70.8 & 76.9 & 70.3 & \textbf{60.4} & 68.0 \\    
     & $p_g$ & S & 89.0 & 93.5 & 83.8 & 0.0 & \textbf{48.7} & 55.6 & 70.5 & \textbf{81.6} & 88.4 & 64.6 & 74.8 & 65.9 & 56.9 & 67.2 \\    
     & $p_l$ & S & 93.8 & \textbf{98.1} & \textbf{85.4} & 0.0 & 48.1 & 55.0 & \textbf{76.8} & 79.8 & \textbf{92.1} & \textbf{79.1} & \textbf{77.6} & \textbf{74.5} & 58.6 & \textbf{70.7} \\
    \hline
    OA-CNNs & $p_s$ & F & 92.9 & \textbf{98.4} & 82.8 & 0.0 & 29.6 & 55.6 & \textbf{84.3} & \textbf{80.3} & 92.5 & 71.4 & \textbf{77.4} & \textbf{73.6} & \textbf{63.7} & 69.4 \\
     & $p_g$ & F & \textbf{93.9} & 98.3 & 84.1 &  0.0  & \textbf{39.5} & 56.6 & 83.0 &  80.1 & 90.6 & 76.3 & 75.3 & 67.0  & 59.6  & 69.6 \\
     & $p_l$ & F & 92.7 & 97.6 & \textbf{84.5} &  0.0 &  35.7 & \textbf{58.9} & 82.8 & 79.2 & \textbf{92.6} & \textbf{78.9} & 75.3 & 72.4 & 62.4 & \textbf{70.3} \\
    \hdashline
    PTv3 & $p_s$ & F & \textbf{95.5} & 98.2 & \textbf{85.4} & 0.0 & \textbf{45.6} & \textbf{62.2} & 76.1 & \textbf{80.4} & 89.9 & 69.9 & \textbf{79.0}  & \textbf{71.8} & \textbf{63.4} & \textbf{70.4} \\
     & $p_g$ & F & 92.2 & 98.2 & 81.3 &  0.0 & 33.5 & 48.7 & \textbf{78.5} & 72.1 & \textbf{90.7} & \textbf{75.1} & 72.4 & 50.4 & 53.3 & 65.1\\
     & $p_l$ & F & 93.2 & \textbf{98.4} & 83.4 & 0.0 & 32.1 & 43.7 & 73.3 & 73.2 & 90.2 & 59.3 & 72.7 & 44.3 & 55.9 & 63.1 \\
     \hdashline
    KPConvD & $p_s$ & F & \textbf{94.6} & 98.1 & 84.2 & 0.0 & 35.1 & \textbf{63.2} & 70.1 & \textbf{81.6} & 91.3 & 72.8 & 77.0 & 76.3 & \textbf{61.3} & 69.7 \\    
     & $p_g$ & F & 89.7 & 95.9 & 84.3 & 0.0 & \textbf{49.8} & 60.8 & 72.0 & \textbf{81.6} & 88.8 & 65.9 & 75.2 & 70.2 & 58.5 & 68.7 \\    
     & $p_l$ & F & 93.7 & \textbf{98.2} & \textbf{84.5} & 0.0 & 48.6 & 59.4 & \textbf{76.8} & 80.8 & \textbf{92.9} & \textbf{82.2} & \textbf{77.6} & \textbf{77.2} & 59.4 & \textbf{71.7} \\
    \bottomrule
 \\
\end{tabular}
\end{table*}

\begin{table*}[]
    \centering
    \footnotesize
    \setlength{\tabcolsep}{4pt}
    \caption{Results on Paris-Lille-3D validation split (Lille2). Evaluation is done on subclouds (S) and on the final full cloud (F).}
    \label{tab:paris_lille_results}
\begin{tabular}{cccccccccccccc}
    \toprule
    \rotatebox{70}{Model} & \rotatebox{70}{Cropping} & \rotatebox{70}{Evaluation} & \rotatebox{70}{Unclassified} & \rotatebox{70}{Ground} & \rotatebox{70}{Building} & \rotatebox{70}{Pole} & \rotatebox{70}{Bollard} & \rotatebox{70}{Trash can} & \rotatebox{70}{Barrier} & \rotatebox{70}{Pedestrian} & \rotatebox{70}{Car} & \rotatebox{70}{Vegetation} & \rotatebox{70}{\textbf{mIoU}} \\
    \midrule
    OA-CNNs & $p_s$ & S & 24.1 & \textbf{96.1} & 97.5 & 62.7 & 68.4 & 67.3 & 48.1 & \textbf{84.8} & 87.7 & \textbf{91.4} & 72.8 \\
     & $p_g$ & S & \textbf{30.2} & 95.6 & \textbf{98.3} & \textbf{66.2} & \textbf{70.3} & 65.4 & \textbf{58.3} & 75.5 & \textbf{92.5} & 89.6 & 74.2 \\
     & $p_l$ & S & 28.2 & 95.6 & 98.0 & 62.8 & 68.6 & \textbf{69.9} & 56.5 & 82.6 & 90.8 & 90.2 & \textbf{74.3} \\
    \hdashline
    PTv3 & $p_s$ & S & \textbf{24.1} & \textbf{96.3} & \textbf{97.3} & \textbf{57.6} & \textbf{61.8} & \textbf{63.1} & 50.5 & \textbf{76.2} & 87.8 & \textbf{91.4} & \textbf{70.6} \\
     & $p_g$ & S & 25.1 & 95.5 & 96.8 & 55.6 & 60.1 & 56.3 & 45.7 & 62.3 & \textbf{89.6} & 88.0 & 67.5 \\
     & $p_l$ & S & 18.6 & 95.8 & 97.0 & 50.2 & 57.5 & 58.1 & \textbf{55.9} & 62.7 & 89.4 & 89.2 & 67.4 \\
    \hdashline
    KPConvD & $p_s$ & S & 30.3 & \textbf{96.2} & 98.0 & \textbf{72.4} & 81.1 & 70.6 & 64.9 & \textbf{86.9} & 90.0  & \textbf{90.8} & 78.1 \\
     & $p_g$ & S & 31.3 & 95.9 & 97.2 & 66.7 & 79.7 & 66.3 & 60.5 & 60.4 & \textbf{93.3} & 87.5 & 73.9 \\
     & $p_l$ & S & \textbf{31.9} & 95.9 & \textbf{98.4} & 70.6 & \textbf{83.9} & \textbf{72.2} & \textbf{73.4} & 80.9 & \textbf{93.3} & 90.5 & \textbf{79.1} \\
    \hline
    OA-CNNs & $p_s$ & F & 27.6 & \textbf{96.4} & 98.0 & 65.6 & \textbf{81.6} & 73.6 & 55.4 & \textbf{95.9} & 90.2 & \textbf{91.3} & 72.8 \\
     & $p_g$ & F & \textbf{35.0} & 95.5 & \textbf{98.6} & \textbf{73.6} & 70.2 & 73.8 & \textbf{70.6} & 90.4 & \textbf{94.1} & 90.0 & 79.2 \\
     & $p_l$ & F & 34.3 & 95.8 & 98.4 & 68.4 & 80.3 & \textbf{77.9} & 70.5 & 92.6 & 91.6 & 89.4 & \textbf{79.9} \\
    \hdashline
    PTv3 & $p_s$ & F & \textbf{30.7} & \textbf{96.6} & \textbf{98.0} & 60.9 & \textbf{74.6} & 69.8 & 67.7 & \textbf{89.2} & 90.8 & \textbf{91.1} & \textbf{76.9} \\
     & $p_g$ & F & 30.2 & 95.5 & 97.5 & \textbf{65.7} & 68.6 & 67.5 & 64.3 & 78.8 & \textbf{94.6} & 88.2 & 75.1 \\
     & $p_l$ & F & 22.4 & 96.3 & 97.7 & 54.1 & 73.8 & \textbf{71.1} & \textbf{68.1} & 82.2 & 92.5 & 89.5 & 74.8 \\
    \hdashline
    KPConvD & $p_s$ & F & 33.8 & \textbf{96.3} & 98.4 & \textbf{80.1} & \textbf{89.8} & \textbf{77.7} & 73.8 & \textbf{94.2} & 91.0  & 90.2 & 82.5 \\
     & $p_g$ & F & 34.4 & 95.4 & 97.2 & 74.0 & 75.9 & 74.4 & 68.8 & 69.1 & \textbf{94.4} & 87.1 & 77.1 \\
     & $p_l$ & F & \textbf{36.3} & 96.1 & \textbf{98.6} & 74.2 & 86.6 & \textbf{77.7} & \textbf{79.4} & 91.3 & 93.9 & \textbf{92.4} & \textbf{82.7} \\
     \bottomrule
\end{tabular}
\end{table*}

The results for S3DIS dataset are shown in Table \ref{tab:s3dis_results}. Although the both alternative methods improve the results for OA-CNNs, the best cropping method is still slightly worse than using split rooms, which achieved an mIoU of 71.1\% in the original work. While the proposed cropping methods decrease the performance of PTv3, it is interesting that the performance gap between split rooms (73.4\% within the original work) and the full scene is much larger than the gap between OA-CNNs and split rooms. Nevertheless, both methods produce remarkable results, proving that strong performance can be achieved without incorporating prior knowledge into the point cloud. Interestingly, the original KPConvD-S model's performance on single rooms is stated as an mIoU of $70.2 \pm 0.8$ (average of 10 tries), which is slightly worse than the linear crop's performance.

The results for Paris-Lille-3D dataset, validated on scene Lille2, are shown in Table \ref{tab:paris_lille_results}. The results obtained using the OA-CNNs and KPConvD models are comparable to those obtained using the SemanticBridge dataset. The larger context of the subcloud improves the model's performance. In this dataset, the impact was higher for the OA-CNNs than for the KPConvD. Interestingly, this is not the case for PTv3. As with the S3DIS data, the model does not benefit from the larger context. Compared to other methods in this benchmark, OA-CNNs with linear cropping achieves a SOTA result on the official test split with an mIoU greater than 84\%. Even though the KPConvD model was better, we did not submit its results since the gap between the spherical and linear cropping ones was smaller. The approach achieves superior results in terms of IoU for the building, pole, and vegetation classes. It has been demonstrated that these classes are considerable in scale and benefit from larger input subclouds.

The results for Toronto3D dataset, validated on scene L002, are shown in Table \ref{tab:toronto_3d_results}. Regardless of the model architecture, linear cropping achieved the best performance for this dataset. A larger receptive field especially impacts the road mark and pole class. In terms of the benchmark, the results for KPConvD are slightly worse than those of the current best-performing models, such as EyeNet++ \cite{yoo_eyenet_2025}, achieving 81.7~\% mIoU. However, we did not optimize the hyperparameters for this specific dataset but rather used the same setup for all outdoor datasets. For this reason, a difference of 1.6~\%, mainly due to the relatively poor performance of the road class, is reasonable.

\begin{table*}[]
    \centering
    \footnotesize
    \setlength{\tabcolsep}{4pt}
    \caption{Results on Toronto3D validation split. Evaluation is done on subclouds (S) and on the final full cloud (F).}
    \label{tab:toronto_3d_results}
\begin{tabular}{cccccccccccc}
\toprule
    \rotatebox{70}{Model} & \rotatebox{70}{Cropping} & \rotatebox{70}{Evaluation} & \rotatebox{70}{Road} & \rotatebox{70}{Road mark} & \rotatebox{70}{Natural} & \rotatebox{70}{Building} & \rotatebox{70}{Utility line} & \rotatebox{70}{Pole} & \rotatebox{70}{Car} & \rotatebox{70}{Fence} & \rotatebox{70}{\textbf{mIoU}} \\
    \midrule
    OA-CNNs & $p_s$ & S & \textbf{87.6} & 43.9 & 93.7 & 87.5 & 83.0 & 66.4 & 84.7 & 30.9 & 72.2 \\
     & $p_g$ & S & \textbf{87.6} & 47.9 & 92.1 & 87.9 & 73.1 & 65.5 & 80.7 & \textbf{36.8} & 71.5 \\
     & $p_l$ & S & 87.5 & \textbf{54.0} & \textbf{94.1} & \textbf{88.8} & \textbf{84.0} & \textbf{75.3} & \textbf{86.0} & 34.0 & \textbf{75.5} \\
    \hdashline
    PTv3 & $p_s$ & S & 87.7 & 4.5 & 92.5 & 87.7 & \textbf{82.2} & 66.9 & \textbf{83.7} & \textbf{33.3} & 67.3  \\
     & $p_g$ & S & \textbf{88.3} & 19.0 & 91.9 & 88.6 & 67.8 & 59.1 & 83.6 & 25.9 & 65.5 \\
     & $p_l$ & S & 88.0 & \textbf{32.4} & \textbf{93.4} & \textbf{88.7} & 81.1 & \textbf{69.8} & 81.7 & 31.9 & \textbf{70.9} \\
    \hdashline 
    KPConvD & $p_s$ & S & 87.2 & 56.1 & \textbf{94.0} & 87.4 & 81.4 & 71.9 & 80.1 & 44.8 & 75.4 \\
     & $p_g$ & S & \textbf{89.4} & 54.4 & 92.2 & \textbf{90.9} & 74.6 & 69.8 & 85.4 & 39.2 & 74.5 \\
     & $p_l$ & S & \textbf{89.4} & \textbf{57.9} & 93.9 & 90.2 & \textbf{84.2} & \textbf{76.5} & \textbf{88.6} & \textbf{46.1} & \textbf{78.4} \\
    \midrule
    OA-CNNs & $p_s$ & F & 89.1 & 44.7 & \textbf{93.4} & \textbf{89.1} & 86.4 & 72.6 & 87.5 & 36.2 & 74.9 \\
     & $p_g$ & F & 88.8 & 39.6 & 90.3 & 87.2 & 79.9 & 71.2 & 82.4 & \textbf{38.1} & 72.2 \\
     & $p_l$ & F &  \textbf{91.9} & \textbf{58.5} & 93.0 & \textbf{89.1} & \textbf{86.8} & \textbf{78.8} & \textbf{87.8} & 35.1 & \textbf{77.7}\\
    \hdashline
    PTv3 & $p_s$ & F & 89.0 & 0.0 & 92.4 & 89.1 & 85.7 & 74.0 & \textbf{89.5} & \textbf{40.4} & 70.0 \\
     & $p_g$ & F & 88.9 & 10.1 & 89.9 & 87.4 & 77.2 & 66.9 & 87.8 & 26.7 & 66.9 \\
     & $p_l$ & F & \textbf{89.6} & \textbf{32.9} & \textbf{93.3} & \textbf{90.4} & \textbf{85.9} & \textbf{76.0} & 86.6 & 37.7 & \textbf{74.1} \\
    \hdashline 
    KPConvD & $p_s$ & F & 88.6 & 57.5 & \textbf{93.6} & 88.4 & 85.8 & 76.7 & 82.8 & 46.2 & 77.5 \\
     & $p_g$ & F & 89.8 & 49.5 & 89.3 & 87.4 & 77.9 & 73.9 & 87.4 & 41.7 & 74.6 \\
     & $p_l$ & F & \textbf{90.7} & \textbf{62.2} & 93.1 & \textbf{90.5} & \textbf{86.6} & \textbf{79.2} & \textbf{89.9} & \textbf{48.5} & \textbf{80.1} \\
    \bottomrule
\end{tabular}
\end{table*}

\section{Discussion}\label{ch:discussion}
The voxel size and sub cloud size are critical factors that have a significant impact on the performance of the chosen architecture. Both parameters depend heavily on the chosen environment. This makes it difficult to find a universal solution. There will be a trade-off regarding resolution when it comes to the size of the objects to recognize and the level of detail. Interestingly, enhancing the outer bound of the cropping method after training the model can improve its performance. Additionally, applying different cropping methods to the trained network resulted in less performance degradation with non-uniform distributions than with uniform ones. The learned patterns for non-uniform distributions seem to generalize more. It seems that changing cropping strategies is a promising data augmentation method, that has the potential to improve point cloud understanding in general.

Although other cropping methods besides spherical ones lead to better model results with discretization (OA-CNNs) and point-based (KPConvD), the attention based method (PTv3) had difficulty in some cases with non-uniform cropping methods. However, it should be noted that attention-based methods primarily rely on large-scale datasets \cite{dosovitskiy_image_2020} or unsupervised pretraining. In all cases, we trained the models from scratch using only the selected dataset. A more complex point distribution, affecting the serialization, paired with this relatively small amount of data could explain why it performed weaker than expected. Furthermore, we selected the hyperparameters based on prior studies leveraging the OA-CNN model. For PTv3, a slightly different configuration might be more suitable. However, the proposed alternative cropping methods outperform the clear-cut methods in all datasets except the indoor one. This can be explained by the fact that the environmental context required for indoor scenes is smaller than that required for outdoor scenes. While indoor semantics are often only related to the room itself, the size of the scene is limited. The larger the objects to be segmented, the greater the increase achieved by the method. For example, pillars or bridge abutments exceeding ten meters in size greatly benefit from this cropping.

So far, only simple numerical methods were used to investigate the performance of the models and their ability to learn these different point distributions. However, since it is demonstrated that the model can learn using crops other than sperical crops with uniform point probability, one could further test other point sampling methods, such as \cite{jiao_weighted_2025}. This way the crop bound could be further increased by still keeping a high-quality geometric consistency.

The results show that the mIoU can be improved by fusing the subclouds into the original full cloud. This is because points are predicted multiple times from different perspectives. This can be considered a form of test-time augmentation, which has already been proven effective for point clouds. However, there may be more possible improvements. For instance, in \cite{kellner_automatic_2026} simple neighborhood voting increased the final result. This raises the question of whether further advanced methods, specifically designed for large-scale point clouds, could increase the quality of the final scene. 

We have only used random or grid selection to extract center points for the subclouds. However, a topic of interest for future work would be to incorporate uncertainty into the selection process. This would allow for more accurate predictions and better-suited center positions in regions where the model is uncertain. Uncertainty can be estimated by considering multiple predictions of nearby points. This approach could further improve the model's performance without requiring significant modifications. Furthermore, the size of the receptive field caused by the model was not taken into account. The findings indicated that increasing the input crop led to enhanced performance; however, this enhancement did eventually reach a saturation point. Given the substantial adaptations to the given architectures that would be required to increase the internal receptive field, we have chosen not to investigate this topic in the present study. However, we intend to do so in future research.

\section{Conclusion}\label{ch:conclusion}
In this work, we propose easy-to-use cropping methods that employ probability functions for large-scale point clouds that cannot be processed all at once. Dividing large point clouds into smaller, more manageable subclouds results in a loss of environmental context. Our proposed approaches are inspired by the simple idea of keeping points dense close to the focus center, while including sparse surrounding context. This provides a broader, global context within the subclouds by avoiding the need to consider a large number of points. Furthermore, the proposed cropping methods do not introduce additional computational complexity or runtime. This approach improves performance, particularly with large-scale outdoor datasets. Without altering the model architectures, we demonstrate that modifying the input enhances the model's performance. Consequently, this approach enhanced the performance of diverse architectures on multiple datasets, achieving new state-of-the-art (SOTA) results on two datasets.

\section*{Acknowledgment}

This research was funded by the German Federal Ministry of Transport (BMV) as part of the mFUND project "RoboTUNN", funding line 2, grant number 19F2280B.

\bibliographystyle{elsarticle-num} 
\bibliography{chapter/refs}



\end{document}